%% file: paper_arxiv_v1.tex
\documentclass{article} %
\usepackage{iclr2026_conference,times}

\input{math_commands.tex}

\usepackage{wrapfig}
\usepackage{hyperref}
\usepackage{url}
\usepackage{graphicx}
\usepackage{booktabs}
\usepackage{tabularx}
\usepackage{multirow}
\usepackage{capt-of}
\usepackage{float}
\usepackage[nolist, nohyperlinks]{acronym}
\usepackage{xcolor}
\usepackage{colortbl}
\usepackage{etoolbox}
\usepackage[
  textsize=scriptsize,
  color=yellow!15,
  bordercolor=yellow!50!black,
  linecolor=yellow!50!black
]{todonotes}

\usepackage{microtype}

\newlength{\summaryoffset}
\makeatletter
\pretocmd{\@todonotes@drawMarginNote}{\vspace*{\summaryoffset}}{}{}
\patchcmd{\@todonotes@drawLineToRightMargin}
  {[yshift=-0.2cm + \@todonotes@tickmarkheight] inText}
  {[yshift=0.2cm + \@todonotes@tickmarkheight] inText}{}{}
\patchcmd{\@todonotes@drawLineToRightMargin}
  {[yshift=-0.2cm] inText}
  {[yshift=0.2cm] inText}{}{}
\makeatother

\makeatletter
\newif\ifshowsummaries
\showsummariesfalse %
\newcommand{\summary}[1]{%
  \leavevmode
  \ifshowsummaries
  \smash{\rlap{%
    \begin{tikzpicture}[overlay]
      \coordinate (summaryanchor) at (0,1.35ex);
      \coordinate (summaryedge) at ([xshift=\linewidth]summaryanchor);
      \draw[yellow!50!black,line width=0.4pt]
        (summaryedge) -- ++(\marginparsep,0) -- ++(0,-\summaryoffset);
      \node[
        anchor=north west,
        draw=yellow!50!black,
        fill=yellow!15,
        line width=0.4pt,
        inner sep=2pt,
        text width=\dimexpr\marginparwidth-6pt\relax,
        align=left,
        font=\scriptsize
      ] at ([xshift=\marginparsep,yshift=-\summaryoffset]summaryedge) {#1};
    \end{tikzpicture}%
  }}%
  \fi
  \summary@gobblepars
}
\def\summary@gobblepars{%
  \@ifnextchar\par{\summary@gobbleone}{\ignorespaces}%
}
\def\summary@gobbleone\par{\summary@gobblepars}
\makeatother

\definecolor{posteriorviolet}{RGB}{112,72,176}
\definecolor{priorblue}{RGB}{38,103,179}
\definecolor{correctionred}{RGB}{190,55,62}
\newcommand{\postc}[1]{{\color{posteriorviolet}#1}}
\newcommand{\priorc}[1]{{\color{priorblue}#1}}
\newcommand{\corrc}[1]{{\color{correctionred}#1}}

\newcommand{\ours}{LSA (ours)}

\newacro{MOS}{Mean Opinion Score}
\newacro{RIR}{room impulse response}
\newacro{NN}{neural network}
\newacro{FLOP}{floating point operations per second}
\newacro{DSM}{denoising score matching}
\newacro{FM}{flow matching}
\newacro{SM}{score matching}
\newacro{SNR}{signal-to-noise ratio}
\newacro{GAN}{generative adversarial network}
\newacro{VAE}{variational autoencoder}
\newacro{NCSN}{noise conditional score network}
\newacro{STFT}{short-time Fourier transform}
\newacro{iSTFT}{inverse short-time Fourier transform}
\newacro{SDE}{stochastic differential equation}
\newacro{EuM}{Euler--Maruyama}
\newacro{ODE}{ordinary differential equation}
\newacro{DNN}{deep neural network}
\newacro{PESQ}{Perceptual Evaluation of Speech Quality}
\newacro{SE}{speech enhancement}
\newacro{BWE}{bandwidth extension}
\newacro{tf}{time--frequency}
\newacro{ELBO}{evidence lower bound}
\newacro{WPE}{weighted prediction error}
\newacro{PSD}{power spectral density}
\newacro{MSE}{mean-squared error}
\newacro{LSTM}{long short-term memory}
\newacro{POLQA}{Perceptual Objective Listening Quality Analysis}
\newacro{SDR}{signal-to-distortion ratio}
\newacro{PSNR}{Peak signal-to-noise ratio}
\newacro{ESTOI}{Extended Short-Term Objective Intelligibility}
\newacro{ELR}{early-to-late reverberation ratio}
\newacro{TCN}{temporal convolutional network}
\newacro{DRR}{direct-to-reverberant ratio}
\newacro{NFE}{number of function evaluations}
\newacro{RTF}{real-time factor}
\newacro{SQA}{speech quality assessment}
\newacro{LSA}{Likelihood Score Approximation}
\newacro{DPS}{Diffusion Posterior Sampling}
\newacro{CSE}{conditional score estimator}
\newacro{MAP}{maximum a posteriori}
\newacro{LSD}{log-spectral distance}

\title{Livin' on a Prior: Likelihood Score \\Approximation for Inverse Problems}

\author{Rostislav Makarov, Tal Peer, Danilo {de Oliveira}, Timo Gerkmann \\ Signal Processing \\ University of Hamburg \\ Hamburg, Germany \\ \texttt{rostislav.makarov@uni-hamburg.de}}

\iclrfinalcopy %
\begin{document}

\maketitle

\begin{abstract}
Generative models have found great success as data-driven methods of solving inverse problems. Two popular approaches work either by combining a pretrained generative prior with a known degradation model, or by training a conditional generative model directly from paired data.
We target a setting that spans both regimes: unknown degradations can be learned from few paired examples, while known degradations  
can be learned from self-generated samples.
We introduce \emph{\ac{LSA}}, a generative framework that keeps a pretrained unconditional model fixed and learns an observation-conditioned model that approximates the likelihood score from paired samples.
Within a conditional stochastic-interpolant framework, \ac{LSA} can be trained in either score or velocity coordinates, independently of the unconditional model's native parameterization, and supports both deterministic and stochastic sampling. 
We further show empirically that the prior model can be swapped post-training while keeping the same LSA model.
Across speech and image inverse problems, \ac{LSA} operates effectively even at roughly $0.01\%$ of the full training dataset. 
On the ImageNet-256 benchmark it achieves competitive or better restoration quality than strong posterior-sampling baselines while requiring up to several orders of magnitude fewer network 
evaluations. 
\end{abstract}

\section{Introduction}

A wide range of problems in audio and imaging revolve around recovering a signal from degraded or incomplete data measurements. 
Examples of such tasks include super-resolution, inpainting, and deblurring in the image domain, as well as denoising, dereverberation, bandwidth extension, and long-audio inpainting in the audio domain~\citep{kawar2022denoising,chung2023diffusion,richter2023speech,moliner2023solving,welker2026realtime}.
Many inverse problems can be described by the observation model
\begin{equation}
    y=\mathcal{A}(x_0)+\eta,
    \label{eq:inverse_problem}
\end{equation}
where an unknown clean sample $x_0$ is corrupted by a measurement/degradation operator $\mathcal{A}$ plus additive measurement noise $\eta$, resulting in the measurement $y$. 
Since the observation model is generally non-invertible and/or corrupted by noise, reconstruction is often ill-posed and recovery relies on a prior over the signals of interest, to provide information that the observed data cannot.

Diffusion-based generative models have become a popular approach to tackling this class of problems. %
From a Bayesian perspective,
the observation process is characterized by a likelihood $p(y\mid x_0)$. Together with a prior distribution $p(x_0)$ over clean data, this likelihood defines the posterior distribution $p(x_0\mid y)\propto p(y\mid x_0)p(x_0)$ which maps the known measurement $y$ to the unknown clean sample $x_0$. The Bayesian decomposition separates reusable knowledge of the clean-data distribution from observation-specific information encoded by the likelihood. In this view, different inverse problems may share the same prior over clean data, which can be represented by an unconditional generative model, while differing only in the likelihood induced by their respective observation process.

When the observation model $\mathcal{A}$ is known, a broad class of posterior-sampling methods combines a pretrained unconditional generative prior with the available measurement model directly during inference~\citep{chung2023diffusion,zhang2025daps,ahmed2025reps,lemercier2025unsupervised}. Such methods require no task-specific paired training data, but rely on access to the observation model to construct observation-dependent guidance during sampling. 

When the observation model is unknown or unavailable, observation-specific information can instead be learned from paired clean--observation examples $(x_0,y)$. Rather than constructing the posterior from an explicit measurement model at sampling time, conditional adaptation methods use these pairs to adapt a pretrained generative model to the inverse problem. 
When only a few paired examples are available, adapting a pretrained prior rather than training a conditional generative model from scratch is significantly more efficient.
This adaptation can be done, for example, with ControlNet~\citep{zhang2023controlnet} or parameter-efficient methods such as LoRA~\citep{hu2022lora}. ControlNet has been shown to remain trainable with limited conditional data, and \citet{xu2025rethinking} use a ControlNet-based conditional score estimator trained on as few as $25$ paired images for inverse problems.

This motivates a natural question: can we keep the clean-data prior fixed and learn only the observation-dependent component, using real paired data when the observation model is unavailable and self-generated pairs when it is known?

In this paper we introduce \emph{Likelihood Score Approximation}, a framework in which a pretrained unconditional generative model is kept fixed and an observation-conditioned likelihood score approximation is learned from paired data. Unknown degradations can be learned from a few examples, while known degradation operators 
allow to work with only self-generated data.
We show that the same framework can be applied to both score- and flow-based generative models and supports deterministic and stochastic sampling with test-time control. We evaluate \ac{LSA} across speech and image inverse problems, including low-data, out-of-domain, and ImageNet $256\!\times\!256$ and $512\!\times\!512$ benchmarks. \Ac{LSA} also offers the possibility of prior swapping: the frozen prior can be replaced after training by another compatible 
generative model without retraining the \ac{LSA}.

\section{Likelihood Score Approximation with Stochastic Interpolants}
\label{sec:methodology}

Our goal is to extend a pretrained unconditional generative model (serving as a data prior) into an observation-conditioned posterior model. We achieve this by keeping the pretrained prior model fixed and approximating a likelihood score from paired data.
In this section, we first review the Gaussian probability path and its score and velocity representations. We then introduce our method, derive the corresponding \ac{LSA} training objectives, and finally describe deterministic and stochastic posterior samplers. Detailed derivations are provided in Appendix~\ref{app:conditional_projection}.

\subsection{Gaussian probability path, score and velocity}
\label{sec:path_background}

The generative models used in this work transform a simple noise distribution into the data distribution. 
We represent this transport through a Gaussian stochastic interpolant~\citep{albergo2023stochastic}, which defines a continuous path of intermediate distributions. 
This formulation places both unconditional and observation-conditioned generation within the same probability-path formulation.
Let $X_0\sim\pi_{\mathrm{data}}$ denote a clean sample from the data distribution and let $\epsilon\sim\mathcal{N}(0,I)$ denote Gaussian noise independent of $X_0$. We consider the one-sided Gaussian stochastic interpolant~\citep{albergo2023stochastic,ma2024sit}
\begin{equation}
    X_t=\alpha_t X_0+\sigma_t\epsilon,
    \qquad t\in[0,1],
    \label{eq:gaussian_path}
\end{equation}
where $t$ denotes the interpolation time and $X_t$ is the corresponding random state.
The schedules $\alpha_t$ and $\sigma_t$ are monotonically decreasing and increasing functions of $t$, respectively, controlling the interpolation between the clean sample and Gaussian noise. 
We choose $\alpha_0=1$, $\sigma_0=0$, and $\alpha_1=0$, such that $X_0\sim\pi_{\mathrm{data}}$ and $X_1=\sigma_1\epsilon\sim\mathcal{N}(0,\sigma_1^2I)$.

\begin{figure}[t]
    \centering
    \includegraphics[width=1.0\linewidth]{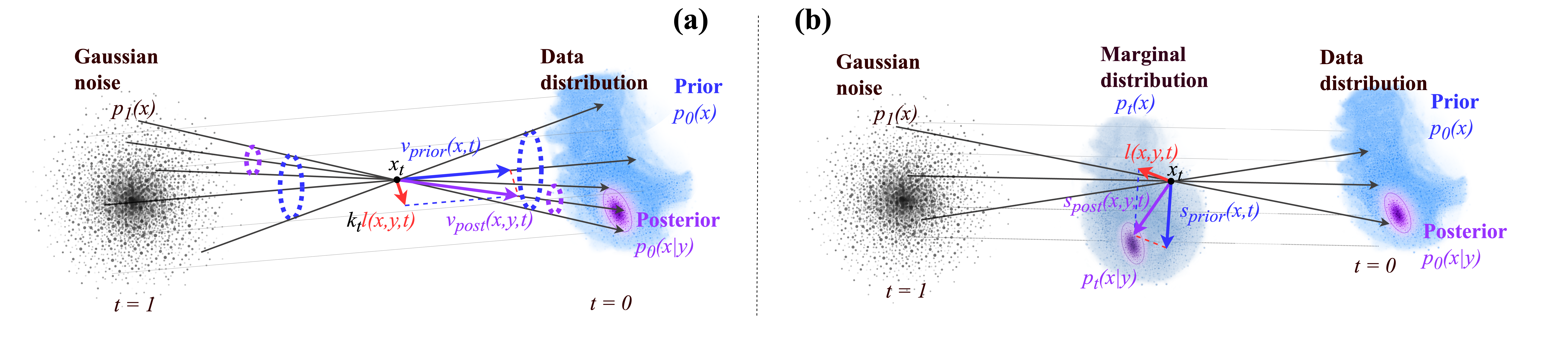}
    \vspace{-.7cm}
    \caption{
    Prior and posterior vector fields along the Gaussian probability path.
    (a) Velocity vectors at the same intermediate state $x_t$, differing by $\kappa_t\ell_t(x,y)$.
    (b) Score vectors for $p_t(x)$ and $p_t(x\mid y)$, differing by $\ell_t(x,y)$.
    }
    \label{fig:scheme}
\end{figure}

Note that, throughout this paper, uppercase letters denote random variables and lowercase letters their realizations. For notational simplicity, we abbreviate $p_{X_t}(x)$, $p_{X_t\mid Y}(x\mid y)$, $p_{Y\mid X_t}(y\mid x)$, and $p_{X_t\mid X_0}(x\mid x_0)$ as $p_t(x)$, $p_t(x\mid y)$, $p_t(y\mid x)$, and $p_{0t}(x\mid x_0)$, respectively.

At each $t$, $X_t$ has marginal density $p_t(x)=\int p_{0t}(x\mid x_0)\pi_{\mathrm{data}}(x_0)\,\mathrm{d}x_0$, where $p_{0t}(x\mid x_0)$ is the conditional density induced by \eqref{eq:gaussian_path}. The marginal density $p_t(x)$ is obtained by integrating out $X_0$. The resulting collection of densities $\{p_t\}_{t\in[0,1]}$ forms a continuous probability path between the data distribution and the Gaussian distribution.

Following the stochastic-interpolant formulation \citep{albergo2023stochastic}, the probability path can be characterized by either of two vector fields: the score field $s(x,t)$, which encodes the geometry of the density $p_t(x)$, and the velocity field $v(x,t)$, which defines its transport over time:
\begin{equation}
    s(x,t)
    :=\nabla_x\log p_t(x)
    =\mathbb{E}\!\left[-\frac{\epsilon}{\sigma_t}\,\middle|\,X_t=x\right],
    \qquad
    v(x,t)
    :=\mathbb{E}\!\left[\dot X_t\,\middle|\,X_t=x\right].
    \label{eq:unconditional_fields}
\end{equation}
Here, $\dot X_t=\dot\alpha_tX_0+\dot\sigma_t\epsilon$ is the time derivative along an interpolant trajectory. 
Since different trajectories can pass through the same state $x$ at time $t$, as illustrated in Fig.~\ref{fig:scheme}(a), $\dot X_t$ is not determined by $X_t=x$, and $v(x,t)$ is its conditional mean.
The conditional-expectation expression for the score holds for $t>0$, where $\sigma_t>0$. 
These definitions follow the standard formulations of score-based diffusion~\citep{song2021scorebased} and flow matching~\citep{lipman2023flowmatching}. We refer to these unconditional vector fields $s(x,t)$ and $v(x,t)$ as the prior vector fields, denoted $s_{\mathrm{prior}}(x,t)$ and $v_{\mathrm{prior}}(x,t)$.

For $t\in(0,1)$, the two fields are related explicitly~\citep{albergo2023stochastic}:
\begin{equation}
    v(x,t)
    =
    \frac{\dot\alpha_t}{\alpha_t}x
    +
    \frac{\sigma_t(\dot\alpha_t\sigma_t-\alpha_t\dot\sigma_t)}
         {\alpha_t}s(x,t),
    \qquad
    s(x,t)
    =
    \frac{\alpha_t v(x,t)-\dot\alpha_t x}
         {\sigma_t(\dot\alpha_t\sigma_t-\alpha_t\dot\sigma_t)}.
    \label{eq:score_velocity_duality}
\end{equation}
These identities allow us to convert between score and velocity representations.

\subsection{Observation-conditioned vector fields}
\label{sec:conditional_fields}

In inverse problems, we seek the posterior distribution conditioned on the observation $y$ introduced in \eqref{eq:inverse_problem}, and accordingly condition the Gaussian probability path in \eqref{eq:gaussian_path} on $y$.
Marginalizing over the clean samples $X_0$ under the clean-data posterior $p(x_0\mid y)$ yields $p_t(x\mid y)=\int p_t(x\mid x_0)p(x_0\mid y)\,\mathrm{d}x_0$, which describes the distribution of the perturbed state $X_t$ at interpolation time $t$.
Applying Bayes' rule then yields the score decomposition:
\begin{equation}
    \underbrace{\postc{\nabla_x\log p_t(x\mid y)}}_{\text{posterior score } \postc{s_{\mathrm{post}}(x,y,t)}}
    =
    \underbrace{\priorc{\nabla_x\log p_t(x)}}_{\text{prior score \priorc{$s_{\mathrm{prior}}(x,t)$}}}
    +
    \underbrace{\corrc{\nabla_x\log p_t(y\mid x)}}_{\text{likelihood score \corrc{$\ell_t(x,y)$}}}
    \label{eq:posterior_decomposition}
\end{equation}
Here, {$\ell_t(x,y)$} denotes the interpolation-time likelihood score.
The interpolation-time likelihood $p_t(y\mid x)$ differs from the clean-data likelihood $p(y\mid x_0)$ because it is conditioned on the perturbed state $X_t=x$ rather than the clean sample $X_0=x_0$, and is therefore generally intractable.
A common line of posterior-sampling methods approximates this intractable term using the known degradation operator $\mathcal{A}$, typically through the clean-data likelihood $p(y\mid x_0)$ evaluated at an estimate of $x_0$~\citep{chung2023diffusion,zhang2025daps,ahmed2025reps,lemercier2025unsupervised}.
In contrast, we propose to learn an approximation of the likelihood score from paired data.

Let $(X_0,Y)\sim\pi_{\mathrm{pair}}$ denote paired clean samples and observations, and draw $\epsilon\sim\mathcal{N}(0,I)$. For a fixed interpolation time $t>0$, we construct $X_t$ using \eqref{eq:gaussian_path}. Conditioning additionally on the observation $Y$ induces the conditional probability path $p_t(x\mid y)$, whose data endpoint is the posterior  $p(x_0\mid y)$. Its corresponding score and velocity vector fields are (cf.~\eqref{eq:unconditional_fields}):
\begin{equation}
    \postc{s_{\mathrm{post}}(x,y,t)}
    =\mathbb{E}\!\left[-\frac{\epsilon}{\sigma_t}\,\middle|\,X_t=x,Y=y\right],
    \qquad
    \postc{v_{\mathrm{post}}(x,y,t)}
    =\mathbb{E}\!\left[\dot X_t\,\middle|\,X_t=x,Y=y\right].
    \label{eq:conditional_fields}
\end{equation}
Importantly, conditioning does not change the sample-wise targets, which remain $-\epsilon/\sigma_t$ for the score and $\dot X_t$ for the velocity. Accordingly, $Y$ enters only as an additional conditioning variable and does not appear in the regression target.
This property was established for conditional score estimation by \citet{batzolis2021conditional} and for conditional velocity fields in flow matching by~\citet{dasgupta2026conditional}. Appendix~\ref{app:conditional_regression} gives the corresponding derivations in our stochastic-interpolant notation.

\subsection{Decomposing posterior vector fields into prior and likelihood terms}
\label{sec:modular_decomposition}

As evident from the score decomposition in \eqref{eq:posterior_decomposition}, the posterior and prior scores differ by the likelihood score. Combining this with the score--velocity relation in \eqref{eq:score_velocity_duality} gives the corresponding difference between their velocity fields:
\begin{equation}
\begin{aligned}
    \postc{s_{\mathrm{post}}(x,y,t)}
    -\priorc{s_{\mathrm{prior}}(x,t)}
    &= \corrc{\nabla_x\log p_t(y\mid x)},\\
    \postc{v_{\mathrm{post}}(x,y,t)}
    -\priorc{v_{\mathrm{prior}}(x,t)}
    &= \kappa_t\,\corrc{\nabla_x\log p_t(y\mid x)},
\end{aligned}
\label{eq:exact_posterior_residual}
\end{equation}
where $\kappa_t=\sigma_t(\dot\alpha_t\sigma_t-\alpha_t\dot\sigma_t)/\alpha_t$. Figure~\ref{fig:scheme} illustrates the corresponding prior and posterior velocity vectors in panel~(a) and score vectors in panel~(b).
Thus, the same likelihood score appears as ${\ell_t(x,y)}$ in score coordinates and as $\kappa_t{\ell_t(x,y)}$ in velocity coordinates.
Besides the score and velocity formulations, the same decomposition can also be expressed in terms of directly predicting a clean-data estimate $\widehat{x}_0$, i.e., with a data prediction loss. This is useful for methods which require an explicit clean-sample estimate during sampling or training. The corresponding derivations are provided in Appendix~\ref{app:field_conversion}.

To turn this decomposition into a learnable model, we keep the pretrained prior fixed and learn the likelihood score approximation separately.
Let $f\in\{s,v\}$ represent either the score ($s$) or velocity ($v$) parameterization of this likelihood term, and let $f_\theta(x,t)$ denote the output of the frozen prior model expressed in the corresponding parameterization (if the prior model uses a different native parameterization, its output must first be converted using \eqref{eq:score_velocity_duality}).
We then introduce the observation-conditioned \ac{LSA} ${\widehat{\ell}_\psi^f(x,y,t)}$ and construct the posterior-field estimator
\begin{equation}
    \postc{\widehat f_{\theta,\psi}(x,y,t)}
    =
    \priorc{f_\theta(x,t)}
    +
    \corrc{\widehat{\ell}_\psi^f(x,y,t)},
    \qquad f\in\{s,v\}.
    \label{eq:prior_residual_model}
\end{equation}
The prior depends only on $(x,t)$, while \ac{LSA} additionally receives the observation $y$. The two models share neither parameters nor intermediate features and are combined only by adding their outputs.

\subsection{Training the Likelihood Score Approximation}
\label{sec:training}

We now turn the additive posterior model into a training objective for \ac{LSA}.
For that, we sample $(X_0,Y)\sim\pi_{\mathrm{pair}}$, $\epsilon\sim\mathcal{N}(0,I)$, and $t\sim\mathcal{U}(t_{\mathrm{eps}},1-t_{\mathrm{eps}})$, where $t_{\mathrm{eps}}>0$ excludes the path endpoints. We construct $X_t$ using \eqref{eq:gaussian_path} and optimize

\begin{equation}
    \mathcal{L}_{\mathrm{LSA}}^{f}(\psi)
    =
    \mathbb{E}_{(X_0,Y),\,t,\,\epsilon}
    \left[
        \lambda_f(t)
        \left\|
            \priorc{\operatorname{sg}\!\left[f_\theta(X_t,t)\right]}
            +
            \corrc{\widehat{\ell}_\psi^f(X_t,Y,t)}
            -
            \postc{T_t^f}
        \right\|_2^2
    \right].
    \label{eq:residual_objectives}
\end{equation}
Here, $\operatorname{sg}[\cdot]$ denotes stop-gradient, so only the \ac{LSA} parameters $\psi$ are optimized, and $\lambda_f(t)>0$ is an optional interpolation time-dependent weight. 
In practice, the expectation in $\mathcal{L}_{\mathrm{LSA}}^f(\psi)$ is approximated by Monte Carlo averaging over sampled minibatches.
We denote by $T_t^f$ the corresponding sample-wise target in the chosen parameterization, with $T_t^s=-\epsilon/\sigma_t$ for the score and $T_t^v=\dot X_t$ for the velocity, as introduced in \secref{sec:conditional_fields}.
By applying \eqref{eq:conditional_fields}, the conditional mean of these targets given $X_t=x$ and $Y=y$ is the corresponding posterior field ${f_{\mathrm{post}}(x,y,t)}$.

Let ${\widehat{\ell}^{f}(x,y,t)^*}$ denote the optimum of the objective above. %
It therefore accounts for the difference between the frozen prior field and the posterior regression optimum. Hence,
\begin{equation}
\begin{aligned}
    \corrc{\widehat{\ell}^{f}(x,y,t)^*}
    &=
    \postc{f_{\mathrm{post}}(x,y,t)}
    -
    \priorc{f_\theta(x,t)}
    \\
    &=
    \underbrace{
        \postc{f_{\mathrm{post}}(x,y,t)}
        -
        \priorc{f_{\mathrm{prior}}(x,t)}
    }_{\corrc{\ell_t^f(x,y)}}
    +
    \underbrace{
        \priorc{f_{\mathrm{prior}}(x,t)}
        -
        \priorc{f_\theta(x,t)}
    }_{\Delta_\theta^f(x,t)} .
\end{aligned}
\label{eq:residual_optima}
\end{equation}
Here, ${\ell_t^f}$ denotes the likelihood score in the chosen parameterization, as defined in \eqref{eq:exact_posterior_residual}.
The term $\Delta_\theta^f$ captures the mismatch between the frozen prior model and the exact prior field. 
Thus, \ac{LSA} can approximate the likelihood score up to the prior-model error $\Delta_\theta^f$.
The corresponding regression derivation and the relation between the score and velocity \ac{LSA} objectives are provided in Appendix~\ref{app:residual_regression}. Figure~\ref{fig:lsa_overview} summarizes the \ac{LSA} training setup. %

\begin{wrapfigure}{r}{0.47\textwidth}
    \centering
    \vspace{-0.5\baselineskip}
    \includegraphics[width=\linewidth]{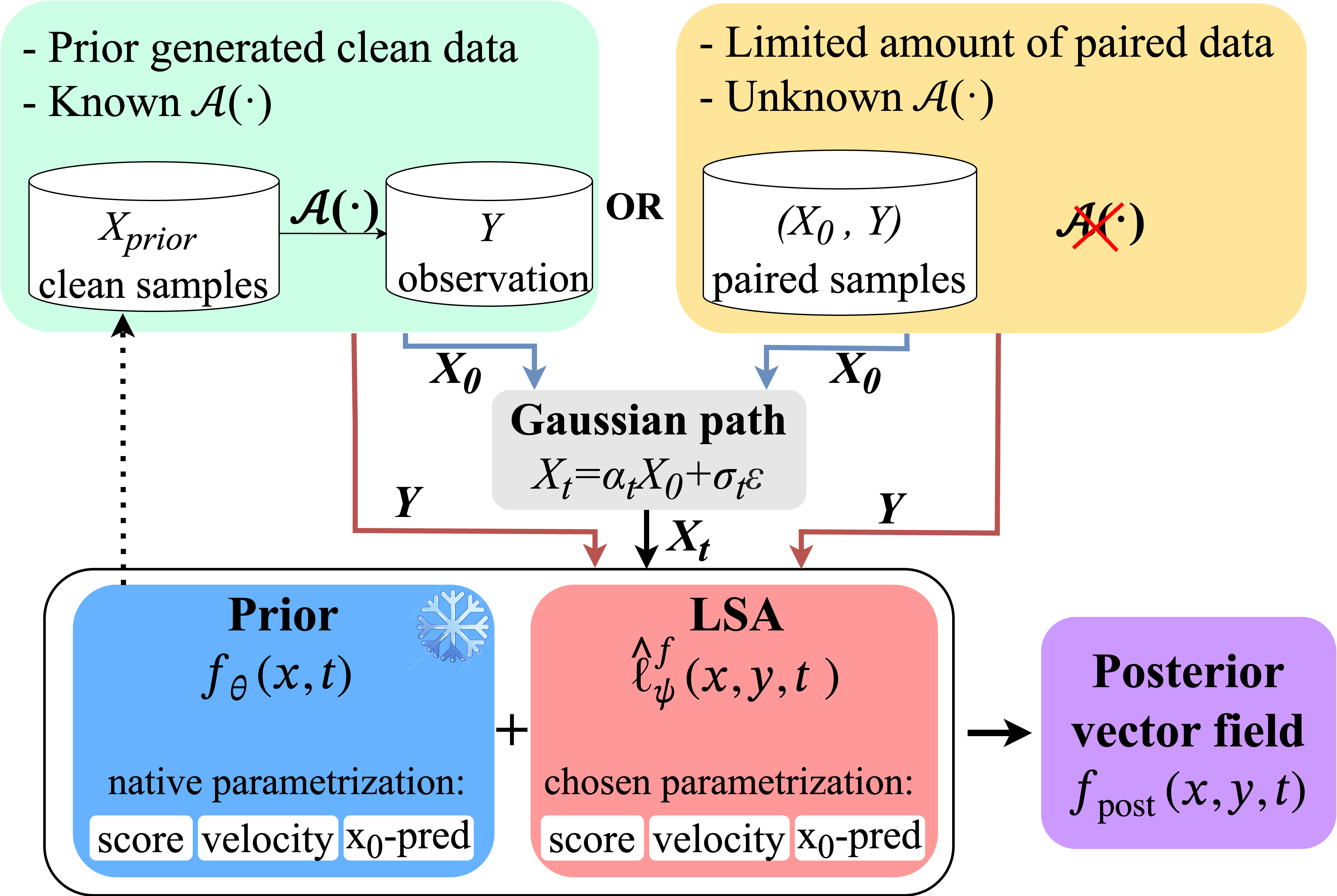}
    \vspace{-1.5\baselineskip}
    \caption{Overview of \ac{LSA}. Training uses either self-generated pairs with a known degradation operator or limited paired-data with an unknown operator. Prior and \ac{LSA} parameterizations can be chosen independently.}
    \label{fig:lsa_overview}
    \vspace{-1.5\baselineskip}
\end{wrapfigure}
\paragraph{Prior swapping.}
The training procedure of \ac{LSA} keeps the prior and \ac{LSA} separate and combines them only through their output vector fields, allowing the prior to be swapped after training without retraining \ac{LSA}. 
For example, if a better prior trained on more data becomes available or a specialized prior better matches the data distribution. The replacement prior must use the same $\alpha_t$ and $\sigma_t$ as in \eqref{eq:gaussian_path}. Since \ac{LSA} is optimized relative to the original frozen prior, swapping the prior does not generally preserve the exact posterior field in ~\eqref{eq:residual_optima}. 
The efficacy of prior swapping is demonstrated in Section~\ref{sec:results_speech} and further analyzed in Appendix~\ref{app:prior_swapping}.

\subsection{Deterministic and stochastic posterior sampling}
\label{sec:sampling}

The frozen prior and \ac{LSA} together define the posterior-field estimator in \eqref{eq:prior_residual_model}. The probability-flow ODE only requires the velocity field, while the stochastic-interpolant SDE additionally uses the score term. When only one representation ${\widehat f_{\theta,\psi}(x,y,t)} \in\{s,v\}$ is produced directly, the other is obtained from \eqref{eq:score_velocity_duality}, without an additional model.
Sampling starts from the Gaussian endpoint $X_1\sim\mathcal{N}(0,\sigma_1^2I)$, and proceeds toward the data endpoint. We use
\begin{equation}
\begin{aligned}
    \text{ODE:}\qquad
    dX_t
    &=
    \postc{\widehat v_{\theta,\psi}(X_t,y,t)}\,dt,\\
    \text{SDE:}\qquad
    dX_t
    &=
    \left[
        \postc{\widehat v_{\theta,\psi}(X_t,y,t)}
        -\frac{1}{2}w_t\postc{\widehat s_{\theta,\psi}(X_t,y,t)}
    \right]dt
    +\sqrt{w_t}\,d\overline W_t,
    \qquad w_t\geq0.
\end{aligned}
\label{eq:posterior_samplers}
\end{equation}
Here, $\overline W_t$ denotes a reverse-time Wiener process and $w_t$ controls the sampler stochasticity. Setting $w_t=0$ recovers the probability-flow ODE. With exact posterior fields, different choices of $w_t$ define different sampling dynamics with the same posterior marginals $p_t(x\mid y)$~\citep{albergo2023stochastic}.

\section{Related work}
\label{sec:related_work}

\paragraph{Learning residual corrections.} %
The closest approach is DEFT~\citep{denker2024deft}, which likewise learns the posterior–prior score correction through a simulation-free denoising objective.
The distinction is therefore 
in a way of representing and using the likelihood correction. In its reported inverse-problem implementations, DEFT conditions the learned \(h\)-transform on prior-derived features such as the unconditional Tweedie estimate and on operator-derived likelihood features; for inpainting, the network additionally receives the observation mask.
Supervised Guidance Training~\citep{baker2026supervised} extends the DEFT principle to infinite-dimensional score-based diffusion models and also considering operator-informed likelihood features.
In contrast, \ac{LSA} receives only $(x,y,t)$, requires neither prior outputs nor operator-derived features, and applies to both score and velocity parameterizations. 
Related flow-matching work also exploit additive conditional structure: CompFlow identifies the conditional--unconditional velocity residual with a noised likelihood gradient but uses it for training-free composition of conditional generators~\citep{miglior2026compflow}.

\paragraph{Low-data conditional adaptation.}
Parameter-efficient adaptation of pretrained diffusion models has also been explored under limited supervision. ControlNet remains trainable with as few as $1$k conditional pairs~\citep{zhang2023controlnet}, and \citet{xu2025rethinking} use a ControlNet-based conditional score estimator trained on only $25$ images for inverse problems. 
Low rank adaptation (LoRA)~\citep{hu2022lora} adapts frozen pretrained models through low-rank weight updates; this strategy has also been applied directly to image restoration, with AdaptSR using LoRA to adapt pretrained super-resolution models to real-world degradations~\citep{korkmaz2025adaptsr}. We therefore include ControlNet and LoRA as representative paired-data adaptation baselines in our low-data experiments.

\begin{figure}[H]
    \centering
    \begin{minipage}[c]{0.25\textwidth}
        \centering
        \textbf{(a)}\\[0.35em]
        \includegraphics[width=.9\linewidth]{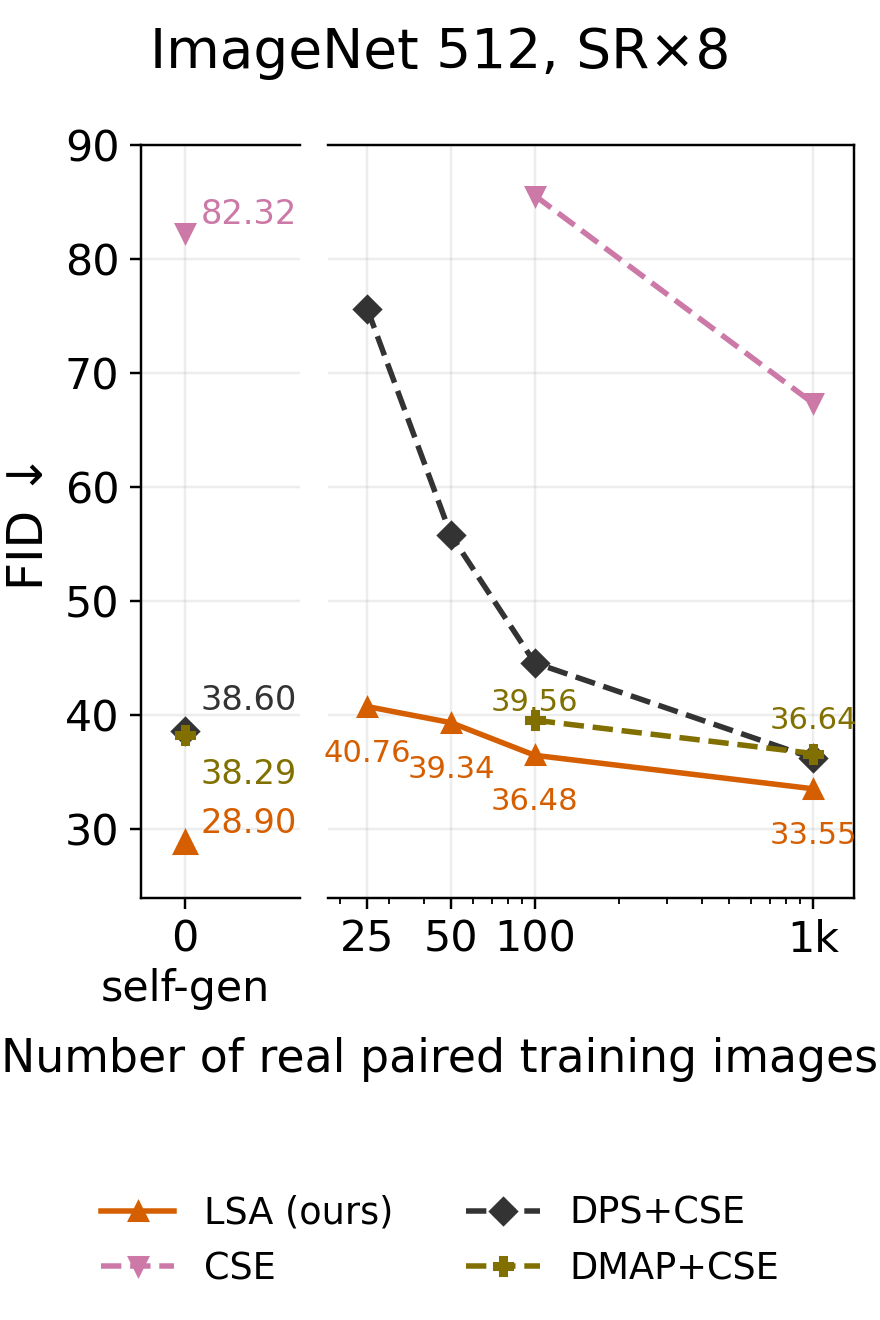}
    \end{minipage}\hfill
    \begin{minipage}[c]{0.73\textwidth}
        \centering
        \textbf{(b)}\\[0.35em]
        \includegraphics[width=.9\linewidth]{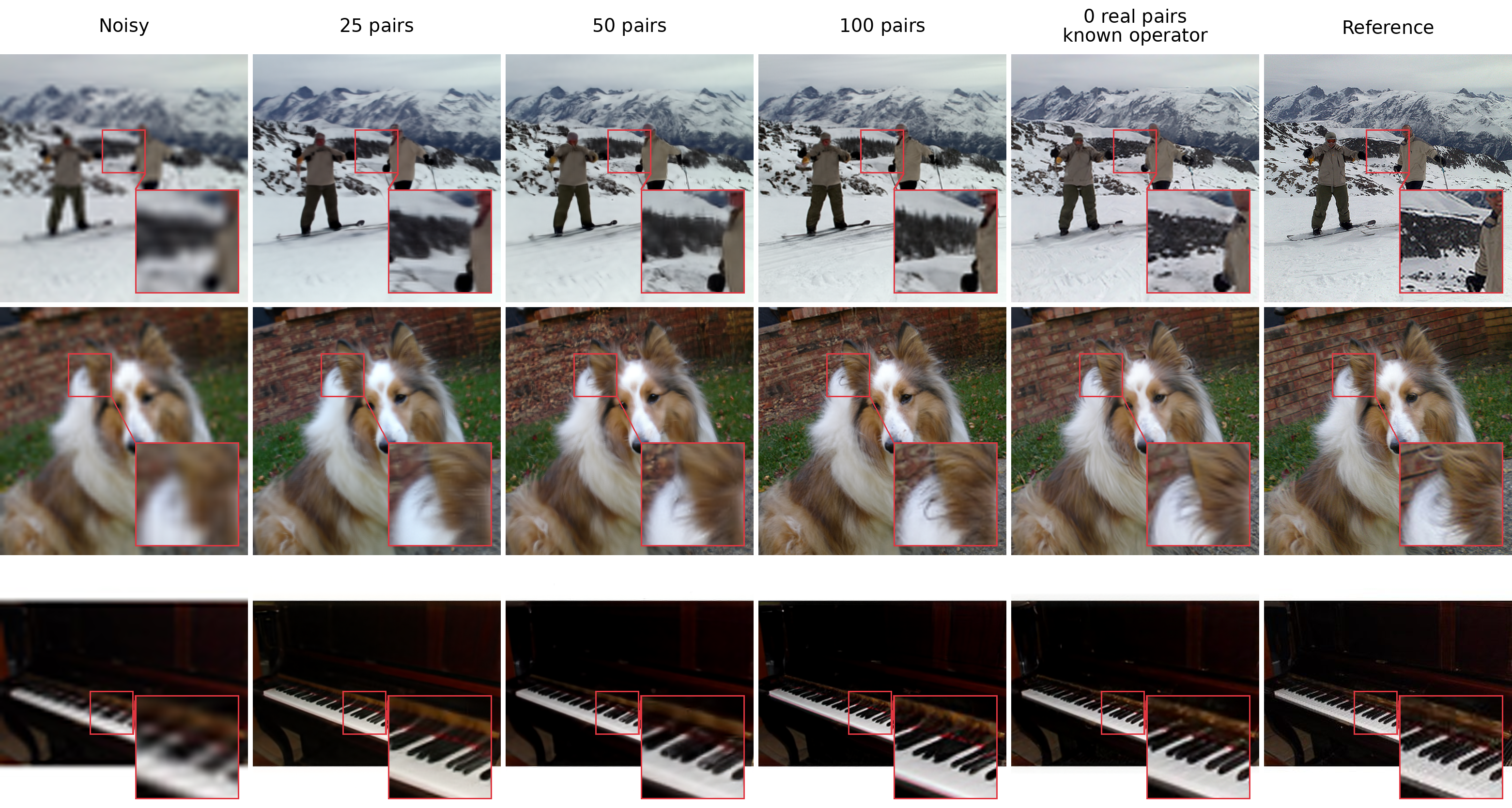}
    \end{minipage}
    \caption{
    ImageNet-$512$ $8\times$ super-resolution across paired-data budgets, following the CSE benchmark~\citep{xu2025rethinking}.
    \textbf{(a)} FID for models trained with 25--1k real pairs, together with the self-generated setting.
    \textbf{(b)} Qualitative results for \ac{LSA}.    
    }   
    \vspace{-1.0\baselineskip}

    \label{fig:imagenet512_srx8_results}
    \label{fig:imagenet512_srx8_data_scaling}
    \label{fig:imagenet512_srx8_qualitative}
\end{figure}

\section{Experimental setup}
\label{sec:experiments}

We evaluate \ac{LSA} in two complementary regimes. First, we study whether a frozen generative prior can be adapted to an unknown observation process from only a small fixed set of paired examples. Second, when the degradation operator $\mathcal{A}$ is known, we show whether real paired data can be removed entirely by training the \ac{LSA} on self-generated samples by the prior model. Across both regimes, we evaluate data efficiency, generality across inverse problems and modalities, and the quality-compute trade-off of the resulting posterior sampler.

We evaluate \ac{LSA} on speech and image restoration tasks. For \emph{speech}, we use clean speech recordings from the EARS dataset \citep{richter24_ears} and consider three inverse problems: speech enhancement, bandwidth extension, and \ac{STFT} phase retrieval. The frozen prior is trained on the full clean training split, whereas \ac{LSA} without access to the degradation operator is trained from fixed paired subsets containing only 1 or 10 minutes of speech $\sim0.16\%$ and $1.6\%$ of EARS training data. The clean--observation pairs are sampled once and then held fixed: during low-data training, neither \ac{LSA} nor the paired baselines access the degradation operator or resample observations online.

For ImageNet-$256$, we follow the DAPS benchmark \citep{zhang2025daps} and evaluate $4\times$ super-resolution, box inpainting and random inpainting, Gaussian, motion, and nonlinear deblurring, and high-dynamic-range reconstruction. In the low-data setting, all paired methods use the same fixed set of 128 clean--observation pairs, $\sim0.01\%$ of ImageNet training data, with a disjoint validation set used for model and sampler selection. Evaluation uses the released 100-image DAPS test subset.

We additionally consider a setting for known degradations. Here, clean training samples are generated from the frozen prior and passed through the known operator to construct self-generated  pairs. The operator is therefore used to create the training observations, but no real paired examples are required. 
Finally, we evaluate $8\times$ super-resolution on ImageNet-$512$ following the low-data protocol of CSE \citep{xu2025rethinking}, utilizing a randomly chosen 25, 50, 100, and 1,000 real paired images together with the self-generated setting.

Across all experiments, we use the linear Gaussian interpolant $\alpha_t=1-t$ and $\sigma_t=c\,t$, where $c=1$ for image models and $c=\sigma_{\max}$ for speech to match the scale of the complex \ac{STFT} representation. Appendix~\ref{app:experimental_details} specializes the score-velocity conversion and posterior ODE/SDE samplers to this path and provides all remaining training, sampling, and hyperparameter-selection details.

\subsection{Models and baselines}
For speech, the frozen prior is an unconditional NCSN++ \citep{song2021scorebased} score model trained on the full clean EARS training set. For the prior-swapping experiments, we additionally train prior models on the Urgent~\citep{Urgent} and VoiceBank~\citep{vbdemand} datasets, as well as an ADM-based~\citep{dhariwal2021diffusion} prior on trained on the EARS.
For images, we use pretrained PixelDiT\footnote{https://huggingface.co/nvidia/PixelDiT-ImageNet} \citep{yu2025pixeldit}  models at $256\times256$ and $512\times512$ resolution and obtain the unconditional velocity field through their learned null condition. The prior remains frozen in all LSA experiments and shares neither parameters nor intermediate features with the \ac{LSA} network.

For \ac{LSA}, only the likelihood-score network is trained on the available paired data, using the objective in~\eqref{eq:residual_objectives}. 
For speech experiments we use the score parameterization, whereas the main image experiments use the velocity parameterization with the frozen PixelDiT velocity prior.
Unless stated otherwise, we use an NCSN++ architecture for the \ac{LSA} network. Architecture and training details are given in Appendix~\ref{app:models_training}.

In the paired low-data setting, we compare primarily against ControlNet and LoRA trained from exactly the same fixed pairs; for speech, we additionally train a direct conditional posterior model by training a single model from scratch with paired data. 
All auxiliary methods are trained to match a posterior vector field directly.
We also evaluate DEFT \citep{denker2024deft} on the tasks for which a matched implementation is available. In the self-generated ImageNet-$256$ setting, we compare against established known-operator posterior samplers under the DAPS protocol. For ImageNet-$512$, we follow the low-data benchmark of \citet{xu2025rethinking} and compare against the main configurations reported in that work.

\input{tables/speech_results_compact.tex}

\subsection{Inference, hyperparameter selection, and evaluation}

By default, we use deterministic posterior sampling with the probability-flow ODE, unless stated otherwise.
For each task, we optimize the inference hyperparameters on held-out validation pairs and keep them fixed for test evaluation. The search covers the number of function evaluations and the two inference refinements described below. 
Sampling hyperparameters are optimized using PESQ/LSD for speech tasks and LPIPS for image tasks.

We additionally use two inference-time refinements to improve restoration quality. 
First, a guidance-like time-dependent scaling factor controls the contribution of the learned \ac{LSA} relative to the frozen prior along the sampling trajectory. This allows us to vary the strength of the \ac{LSA} model, analogously to guidance mechanisms used in diffusion models~\citep{dhariwal2021diffusion,ho2022classifierfree}.
Second, we use a short restart-based refinement, motivated by restart-based generative and posterior sampling~\citep{ahmed2025reps,xu2023restart,zhang2025daps}. After obtaining a restored estimate at the end point of the conditional trajectory, we perform a forward noising step to reintroduce noise at an intermediate point of the Gaussian path. From this restarted state, we integrate the conditional trajectory to the data endpoint again using the same observation $y$. Since the restart uses a newly sampled noise realization, the refinement follows a different conditional trajectory. By default, we perform a single restart unless stated otherwise.
The definitions, interpretation, and hyperparameter optimization are detailed in Appendix~\ref{app:inference_refinements}. All additional evaluations introduced by refinement are included in the reported NFE. %

For image restoration, we report the default set used in inverse problems benchmarks: \ac{PSNR}, SSIM~\cite{zhou2004image}, LPIPS~\cite{zhang2018lpips}, and FID~\cite{heusel2017gans}. For speech, the main tables show four complementary measures per task: PESQ~\cite{rix2001pesq}, DNN-based MOS prediction (as an average of SCOREQ~\cite{ragano2024scoreq}, DistillMOS~\cite{stahl2025distillation}, and NISQA~\cite{mittag2021nisqa}), FAD~\cite{kilgour2019frechet, gui2024adapting}, and either SI-SDR~\cite{leroux2019sdr} for speech enhancement or \ac{LSD} for bandwidth extension and phase retrieval.

\input{tables/image_no_real_pairs_results.tex}

\section{Results}
\label{sec:results}
\renewcommand*{\theHtable}{results.\arabic{table}}
\subsection{Speech inverse problems}
\label{sec:results_speech}

Evaluation results on the three speech restoration tasks are shown in Table~\ref{tab:speech_main_results}. In the speech enhancement task, all models were trained on the EARS-WHAM-v2 dataset \citep{richter24_ears}. The proposed \ac{LSA} approach achieves considerably better results in the low-data scenario (1 min / 10 min of paired data) compared to both ControlNet and direct posterior training, especially w.r.t. FAD and DNN-based MOS scores. In the full-data regime, \ac{LSA} still achieves competitive results, although the improvement upon direct posterior training is less pronounced in this case. Out-of-domain evaluation on the VoiceBank-DEMAND test set shows a certain degradation in performance. This degradation can be mitigated using prior-swapping: either by using a prior trained on a larger amount of data (U) or a prior trained on a smaller dataset which is more similar to the test data (VB). In both cases, only the pretrained prior model is swapped, while the \ac{LSA} model is left unchanged (trained on the in-domain EARS-WHAM-v2).

Bandwidth extension shows the same low-data trend. With only 1 or 10 minutes of paired data, \ac{LSA} achieves better results than both ControlNet and direct posterior training across all reported metrics. Figure~\ref{fig:bwe_qualitative_data_scaling} in Appendix~\ref{app:bwe_qualitative_ablation} provides a qualitative spectrogram comparison across paired-data budgets. 
In the case of \ac{STFT} phase retrieval, \ac{LSA} achieves a reconstruction quality that is superior to other full-data approaches, even when using very limited paired data (1 min). In the self-generated data scenario with known corruption operator, \ac{LSA} can harness the data generated by the simple forward operator and achieves an almost perfect reconstruction in terms of PESQ (4.55) and FAD (0.011).

\subsection{Image inverse problems}
Results for the tasks included in the DAPS benchmark \citep{zhang2025daps} are presented in Table~\ref{tab:image_no_real_pairs_results}. \ac{LSA} is compared against DAPS and, where applicable, against an additional current DPS-based baseline (chosen by best performance in terms of FID on each task).

These include DMAP~\citep{xu2025rethinking}, RePS~\citep{ahmed2025reps}, RED-diff~\citep{mardani2024variational}, and MAS~\citep{zhang2025measurement}.
Compared to the competing approaches, \ac{LSA} achieves considerably lower FID scores, as well as modestly better LPIPS and SSIM values and comparable PSNR. Note that all methods considered are prior-based and do not have access to paired data. However, the forward corruption operator is always known. Unlike the reported baselines, which are all based on the DPS paradigm and backpropagate through the forward operator, the conditional model obtained by \ac{LSA} requires additional training using self-generated paired data derived from the known operator. Nevertheless, besides the improved reconstruction quality, this also allows for a drastic gain in inference efficiency: \ac{LSA} requires fewer model evaluations; in some cases several orders of magnitude fewer than DPS-based approaches. Representative samples are shown in Figures~\ref{fig:imagenet256_qualitative_blurs}--\ref{fig:imagenet256_qualitative_other_operators} in Appendix~\ref{app:imagenet256_low_data_results}.

\begin{figure}[t]
    \centering
    \includegraphics[width=\linewidth]{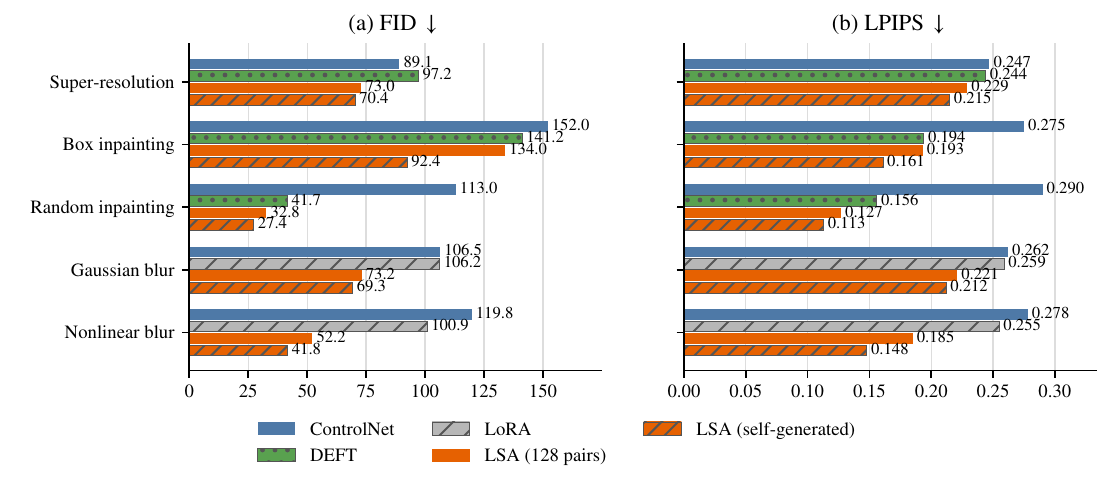}
    \vspace{-.9cm}
    \caption{ImageNet-$256$ restoration with 128 real pairs. All paired methods are trained on the same fixed 128 pairs, with DEFT~\citep{denker2024deft} included where available. For reference, we also show \ac{LSA} trained on self-generated prior samples paired with observations from the known degradation operator.}
    \label{fig:image_low_data_primary_metrics}
\end{figure}

\textbf{ImageNet-256, low-data regime.}
In Figure~\ref{fig:image_low_data_primary_metrics}, we report FID and LPIPS scores on several image reconstruction tasks. \ac{LSA} is compared here against ControlNet, LoRA and DEFT, all using 128 samples of paired data. 
In addition, self-generated results for \ac{LSA} are included as a reference for the known-operator regime (see Table~\ref{tab:image_no_real_pairs_results}). 
\ac{LSA} 
achieves strong reconstruction performance across all tasks
and surpasses all low-data baselines in both metrics, in some cases approaching its self-generated known-operator counterpart.
Full metrics for this comparison are presented in Table~\ref{tab:image_low_data_results} in Appendix~\ref{app:imagenet256_low_data_results}.

\textbf{ImageNet-512 8$\times$ super-resolution.} 
For the specific task of 8$\times$ super-resolution on ImageNet-512 data, we evaluate \ac{LSA} against CSE-based methods \citep{xu2025rethinking} for both the low-data regime and the self-generated known-operator case. 
In terms of FID (Figure~\ref{fig:imagenet512_srx8_results}a), \ac{LSA} performs better than these methods on the self-generated data known-operator case and consistently well on the low-data regime, only requiring 100 data pairs for better reconstruction than the best-performing CSE method (DMAP+CSE) achieves with 1000 pairs.  
For a qualitative evaluation, we provide reconstructed images in Figure~\ref{fig:imagenet512_srx8_results}b for different amounts of paired data. Full comparison with CSE-based methods~\citep{xu2025rethinking} can be found in Table~\ref{tab:imagenet512_srx8_low_data} in Appendix~\ref{app:imagenet512_srx8_results}.

\section{Conclusion}
We introduced \ac{LSA}, a generative framework for inverse problems that keeps a pretrained unconditional prior model fixed and approximates the observation-dependent likelihood score from paired data. Across speech and image restoration tasks, \ac{LSA} remains effective in data-limited regimes, including settings with only minutes of paired speech or 25 paired ImageNet examples. 
When the degradation operator is known, the same framework can be trained from self-generated pairs and substantially reduce inference cost relative to diffusion posterior-sampling approaches.
The same formulation applies to score and velocity formulations, supports deterministic and stochastic posterior sampling, and allows the prior model to be replaced at inference time without retraining \ac{LSA}.

\clearpage

\subsubsection*{Acknowledgments}
Funded by the Deutsche Forschungsgemeinschaft (DFG, German Research Foundation) -- 545210893, 498394658. The authors gratefully acknowledge the scientific support and HPC resources provided by the Erlangen National High Performance Computing Center (NHR@FAU) of the Friedrich-Alexander-Universität Erlangen-Nürnberg (FAU) under the NHR project f102ac. NHR funding is provided by federal and Bavarian state authorities. NHR@FAU hardware is partially funded by the German Research Foundation (DFG) -- 440719683. 

The authors thank Bunlong Lay for insightful scientific discussions and Karim Djemai for his help with figure design.

\subsection*{AI use statement}

Generative AI tools were used to assist with language editing and local rewriting, literature retrieval and discovery, and limited coding support for debugging.

Generative AI tools were not used to formulate the research question, develop the core method or conceptual framework, formulate mathematical claims or derivations, provide proof arguments, propose hypotheses, design the experimental methodology, generate or process datasets, or interpret the reported results. The scientific ideas, methodology, mathematical development, experimental design, and conclusions were developed by the authors.

All AI-assisted text, literature suggestions, and code were reviewed and verified by the authors. References suggested by AI-assisted search were read by the authors, and AI-assisted code was tested before use. The authors take full responsibility for the final content of this work.

\bibliography{iclr2026_conference}
\bibliographystyle{iclr2026_conference}

\appendix

\newpage

\section{Conditional vector fields and \ac{LSA} regression}
\label{app:conditional_projection}

This appendix collects the derivations underlying the conditional regression, score--velocity conversion and \ac{LSA} objectives.

\subsection{Conditional regression from paired samples}
\label{app:conditional_regression}

This subsection justifies the conditional field identities in \eqref{eq:conditional_fields}. For paired samples $(X_0,Y)\sim\pi_{\mathrm{pair}}$, conditioning on a fixed observation $Y=y$ changes the clean endpoint distribution to $p(x_0\mid y)$, while the Gaussian interpolant and its sample-wise score and velocity targets remain unchanged. We show below that these targets correspond to the posterior score and velocity fields of the conditional probability path $p_t(x\mid y)$.

For the Gaussian interpolant in \eqref{eq:gaussian_path}, the conditional marginal at time $t$ is
\begin{equation}
    p_t(x\mid y)
    =
    \int p_{0t}(x\mid x_0)p(x_0\mid y)\,dx_0,
    \qquad
    p_{0t}(x\mid x_0)
    =
    \mathcal{N}(x;\alpha_t x_0,\sigma_t^2 I).
    \label{eq:conditional_path_marginal}
\end{equation}

Differentiating the logarithm of the conditional marginal gives
\begin{align}
    s_{\mathrm{post}}(x,y,t)
    &:=
    \nabla_x\log p_t(x\mid y)
    \label{eq:conditional_score_step1}\\
    &=
    \frac{1}{p_t(x\mid y)}
    \int p(x_0\mid y)\nabla_x p_t(x\mid x_0,y)\,dx_0
    \label{eq:conditional_score_step2}\\
    &=
    \int
    \frac{p_t(x\mid x_0,y)p(x_0\mid y)}
         {p_t(x\mid y)}
    \nabla_x\log p_t(x\mid x_0,y)\,dx_0
    \label{eq:conditional_score_step3}\\
    &=
    \int
    p(x_0\mid x,y)
    \nabla_x\log p_t(x\mid x_0,y)\,dx_0
    \label{eq:conditional_score_step4}\\
    &=
    \int
    p(x_0\mid x,y)
    \nabla_x\log p_{0t}(x\mid x_0)\,dx_0
    \label{eq:conditional_score_step5}\\
    &=
    \mathbb{E}\!\left[
        \nabla_x\log p_{0t}(x\mid x_0)
        \,\middle|\,
        X_t=x,Y=y
    \right].
    \label{eq:conditional_score_marginalization}
\end{align}

The transition from line~(\ref{eq:conditional_score_step4}) to line~(\ref{eq:conditional_score_step5}) follows from the construction of the Gaussian interpolant. After conditioning on $X_0=x_0$, the perturbed state is generated as $X_t=\alpha_t x_0+\sigma_t\epsilon$, where $\epsilon$ is independent of $Y$. 
Since $\epsilon$ is sampled independently of $(X_0,Y)$, we have $X_t\perp Y\mid X_0$ and

\begin{equation}
    p_t(x\mid x_0,y)
    =
    p_{0t}(x\mid x_0).
\end{equation}

For the Gaussian transition,
\begin{equation}
    \nabla_x\log p_{0t}(x\mid x_0)
    =
    -\frac{x-\alpha_t x_0}{\sigma_t^2}
    =
    -\frac{\epsilon}{\sigma_t},
    \label{eq:gaussian_transition_score}
\end{equation}
and therefore
\begin{equation}
    s_{\mathrm{post}}(x,y,t)
    =
    \mathbb{E}\!\left[
        -\frac{\epsilon}{\sigma_t}
        \,\middle|\,
        X_t=x,Y=y
    \right].
    \label{eq:conditional_score_projection}
\end{equation}

For flow matching, the sample-wise target remains
\begin{equation}
T_t^{\mathrm{v}}
=
\dot X_t
=
\dot\alpha_tX_0+\dot\sigma_t\epsilon.
\label{eq:conditional_velocity_target}
\end{equation}
To obtain the corresponding posterior velocity field, consider a fixed observation $Y=y$. Conditioning on $Y=y$ restricts the clean endpoint to $X_0\sim p(x_0\mid y)$, while the interpolant $X_t=\alpha_tX_0+\sigma_t\epsilon$ and its sample-wise velocity $\dot X_t$ remain unchanged. The resulting conditional probability path is $p_t(x\mid y)$, whose data endpoint is the posterior $p(x_0\mid y)$. Applying the same stochastic-interpolant velocity identity as in \eqref{eq:unconditional_fields} to this conditional distribution therefore gives
\begin{equation}
    v_{\mathrm{post}}(x,y,t)
    =
    \mathbb{E}\!\left[
        \dot X_t
        \,\middle|\,
        X_t=x,Y=y
    \right].
    \label{eq:conditional_velocity_projection}
\end{equation}
Thus, $v_{\mathrm{post}}$ is the velocity field associated with the conditional probability path $p_t(x\mid y)$: it averages the same sample-wise velocities $\dot X_t$, but only over trajectories compatible with both $X_t=x$ and $Y=y$. For a derivation showing that this velocity field transports the corresponding conditional probability path, see \citet[Proposition~2.1 and Section~2.2]{dasgupta2026conditional}.

\subsection{Score-velocity conversion and posterior-prior decomposition}
\label{app:field_conversion}

Let $\mathcal{C}=X_t=x$ for the prior field and $\mathcal{C}=(X_t=x,Y=y)$ for the posterior field. In either case, the Gaussian path gives
\begin{equation}
    x = \alpha_t \mathbb{E}[X_0 \mid \mathcal{C}]
        + \sigma_t \mathbb{E}[\epsilon \mid \mathcal{C}],
    \qquad
    v(x,t) = \dot{\alpha}_t \mathbb{E}[X_0 \mid \mathcal{C}]
        + \dot{\sigma}_t \mathbb{E}[\epsilon \mid \mathcal{C}],
\end{equation}
with
\[
    \mathbb{E}[\epsilon \mid \mathcal{C}] = -\sigma_t s(x,t).
\]
Eliminating the conditional clean mean gives the score--velocity relation in \eqref{eq:score_velocity_duality} for either choice of $\mathcal{C}$. Hence,
\begin{equation}
\begin{aligned}
    v_{\mathrm{prior}}(x,t)
        &= \frac{\dot{\alpha}_t}{\alpha_t} x
        + \kappa_t s_{\mathrm{prior}}(x,t), \\
    v_{\mathrm{post}}(x,y,t)
        &= \frac{\dot{\alpha}_t}{\alpha_t} x
        + \kappa_t s_{\mathrm{post}}(x,y,t),
\end{aligned}
\qquad
    \kappa_t :=
    \frac{\sigma_t \bigl(\dot{\alpha}_t \sigma_t - \alpha_t \dot{\sigma}_t\bigr)}
         {\alpha_t}.
\label{eq:prior_posterior_score_velocity_app}
\end{equation}

By Bayes' rule,
\begin{equation}
    s_{\mathrm{post}}(x,y,t) - s_{\mathrm{prior}}(x,t)
        = \nabla_x \log p_t(y \mid x)
        =: \ell_t(x,y).
    \label{eq:bayes_score_residual_app}
\end{equation}
Subtracting the two instances of \eqref{eq:prior_posterior_score_velocity_app} cancels the common term $\frac{\dot{\alpha}_t}{\alpha_t}x$, giving
\begin{equation}
\begin{aligned}
    v_{\mathrm{post}}(x,y,t) - v_{\mathrm{prior}}(x,t)
        &= \kappa_t \left(
            s_{\mathrm{post}}(x,y,t) - s_{\mathrm{prior}}(x,t)
        \right) \\
        &= \kappa_t \ell_t(x,y).
\end{aligned}
\label{eq:posterior_velocity_residual_app}
\end{equation}
Thus, the posterior-prior difference carries the same likelihood score in score and velocity coordinates, but scaled by $\kappa_t$ for the velocity. The same conditional--unconditional velocity residual was derived by \citet{miglior2026compflow} for a specific Gaussian probability path.

\paragraph{Data prediction formulation.}

The same conditional probability path can also be represented through the conditional mean of the clean-data $X_0$. We define the prior and posterior data-prediction fields as
\begin{equation}
    d_{\mathrm{prior}}(x,t)
    :=
    \mathbb{E}[X_0\mid X_t=x],
    \qquad
    d_{\mathrm{post}}(x,y,t)
    :=
    \mathbb{E}[X_0\mid X_t=x,Y=y].
\end{equation}

Unlike the velocity field, these quantities do not themselves define the transport of the probability path. For the Gaussian interpolant, however, they are exactly related to the corresponding score and velocity fields. For either the prior or posterior conditioning, using \eqref{eq:gaussian_path} gives
\begin{equation}
    d(x,t)
    =
    \frac{x+\sigma_t^2 s(x,t)}{\alpha_t}
    =
    \frac{\sigma_t v(x,t)-\dot{\sigma}_t x}
    {\dot{\alpha}_t\sigma_t-\alpha_t\dot{\sigma}_t}.
    \label{eq:data_prediction_conversion_app}
\end{equation}
These relations correspond to the one-sided stochastic-interpolant identities in \citet[Section~4.4, Eqs.~(4.16)--(4.18)]{albergo2023stochastic}.
Thus, the same data prediction can be obtained from either the score or velocity field.

Again, applying \eqref{eq:data_prediction_conversion_app} to the prior and posterior fields and subtracting gives
\begin{equation}
    \begin{aligned}
    d_{\mathrm{post}}(x,y,t)-d_{\mathrm{prior}}(x,t)
    &=
    \frac{\sigma_t^2}{\alpha_t}
    \left(
    s_{\mathrm{post}}(x,y,t)-s_{\mathrm{prior}}(x,t)
    \right) \\
    &=
    \frac{\sigma_t^2}{\alpha_t}\ell_t(x,y).
    \end{aligned}
    \label{eq:posterior_data_residual_app}
\end{equation}
Thus, as in the velocity posterior-prior difference \eqref{eq:posterior_velocity_residual_app}, the posterior-prior difference in data-prediction coordinates is proportional to the same likelihood score $\ell_t(x,y)$, up to a known path-dependent factor $\sigma_t^2/\alpha_t$.

\subsection{\ac{LSA} regression and parameterization equivalence}
\label{app:residual_regression}

Choose the \ac{LSA} parameterization $f\in\{s,v\}$ and express the frozen prior output as $f_\theta$ using \eqref{eq:unconditional_fields}. The assembled predictor is
\begin{equation}
    \widehat f_{\theta,\psi}(x,y,t)
    =
    f_\theta(x,t)
    +
    \widehat{\ell}_\psi^f(x,y,t)
    \label{eq:assembled_residual_predictor_app}
\end{equation}
Here, $f_\theta$ remains fixed during training, regardless of the prior's native parameterization. 
We consider squared-error regression of the assembled predictor toward the sample-wise target $T_t^f$, while keeping $f_\theta$ fixed. At the population level, assuming an unrestricted \ac{LSA} predictor, i.e., one that can represent any square-integrable function of $(x,y,t)$, optimizing the \ac{LSA} model is equivalent to optimizing the assembled predictor.
By the standard squared-error regression identity, the population optimum of the assembled predictor is
\begin{equation}
    \widehat f_{\theta,\psi}^{\,*}(x,y,t)
    =
    \mathbb{E}\!\left[
        T_t^f
        \,\middle|\,
        X_t=x,Y=y
    \right]
    =
    f_{\mathrm{post}}(x,y,t)
    \label{eq:assembled_population_optimum_app}
\end{equation}
Thus, the assembled model has the same population optimum as ordinary conditional score or flow regression.

Since $f_\theta$ is fixed, the optimal likelihood score approximation is obtained by subtracting the frozen field:
\begin{equation}
    \widehat\ell_f^*(x,y,t)
    =
    f_{\mathrm{post}}(x,y,t)
    -
    f_\theta(x,t)
    \label{eq:residual_population_optimum_app}
\end{equation}
Thus, at the unrestricted population optimum, \ac{LSA} recovers the difference between the posterior field and the frozen prior model.

A time-weighted squared-error objective that achieves this population optimum for both parameterizations is
\begin{equation}
    \mathcal{L}_{\mathrm{LSA}}^{f}(\psi)
    =
    \mathbb{E}_{\substack{
        (X_0,Y)\sim\pi_{\mathrm{pair}}\\
        t\sim\mathcal{U}(t_{\mathrm{eps}},1-t_{\mathrm{eps}})\\
        \epsilon\sim\mathcal{N}(0,I)
    }}
    \left[
        \lambda_f(t)
        \left\|
            f_\theta(X_t,t)
            +
            \widehat{\ell}_\psi^{f}(X_t,Y,t)
            -
            T_t^{f}
        \right\|_2^2
    \right]
    \label{eq:generic_residual_objective_app}
\end{equation}
Since $\lambda_f(t)>0$ depends only on $t$, the weighting does not change the pointwise population optimum derived above.
The prior output, likelihood score approximation, and sample-wise target are all expressed in the chosen parameterization $f$. 

The two forms are connected by the same affine score--velocity relation derived above. Using $X_t=\alpha_tX_0+\sigma_t\epsilon$ to eliminate $X_0$ gives
\begin{equation}
    T_t^{v}
    =
    \frac{\dot\alpha_t}{\alpha_t}X_t
    +
    \kappa_t T_t^{s}.
    \label{eq:pathwise_target_conversion_app}
\end{equation}
To compare the two objectives, we express the same frozen prior in score and velocity coordinates and use $\widehat{\ell}_\psi^v=\kappa_t \widehat{\ell}_\psi^s$. The assembled fields then satisfy:
\begin{equation}
    \widehat v_{\theta,\psi}
    =
    \frac{\dot\alpha_t}{\alpha_t}X_t
    +
    \kappa_t\widehat s_{\theta,\psi}.
    \label{eq:assembled_field_conversion_app}
\end{equation}
Subtracting the corresponding targets cancels the common affine term, yielding
\begin{equation}
    \left\|
        \widehat v_{\theta,\psi}-T_t^{v}
    \right\|_2^2
    =
    \kappa_t^2
    \left\|
        \widehat s_{\theta,\psi}-T_t^{s}
    \right\|_2^2.
    \label{eq:residual_loss_equivalence_app}
\end{equation}
For these compatible representations, the unweighted losses at fixed time differ by $\kappa_t^2$, and their population optima correspond under the same conversion. Choosing $\lambda_s(t)=\kappa_t^2\lambda_v(t)$ makes the time-weighted objectives identical.

\paragraph{Data-prediction parameterization.}
The same residual-regression construction applies in data-prediction
coordinates, using the sample-wise target $T_t^d=X_0$.
Accordingly, the corresponding \ac{LSA} objective is obtained from
\eqref{eq:generic_residual_objective_app} by replacing
$f_\theta$, $\widehat{\ell}_\psi^f$, and $T_t^f$ with
$d_\theta$, $\widehat{\ell}_\psi^d$, and $X_0$, respectively.

\section{Experimental details}
\label{app:experimental_details}

\subsection{Linear Gaussian probability path}
\label{app:linear_gaussian_path}

Throughout the experiments, we use the linear Gaussian path
\begin{equation}
    X_t = (1-t)X_0 + ct\epsilon,
    \qquad
    \epsilon\sim\mathcal{N}(0,I),
    \qquad t\in[0,1],
    \label{eq:linear_gaussian_path}
\end{equation}
corresponding to $\alpha_t=1-t$ and $\sigma_t=ct$.
We use $c=1$ for image experiments and $c=\sigma_{\max}=0.5$ for speech, where the latter accounts for the scale of the compressed complex-\ac{STFT} representation.

Substituting these schedules into Eq.~\eqref{eq:score_velocity_duality} gives the direct conversion between velocity and score fields:
\begin{equation}
    v(x,t)
    =
    -\frac{x+c^2t\,s(x,t)}{1-t},
    \qquad
    s(x,t)
    =
    -\frac{(1-t)v(x,t)+x}{c^2t}.
    \label{eq:linear_score_velocity}
\end{equation}
Thus, trained score or velocity predictions can be converted analytically into the representation required by the sampler without an additional network.

For this path, the sample-wise \ac{LSA} training targets from Eq.~\eqref{eq:residual_objectives} reduce to
\begin{equation}
    T_t^s=-\frac{\epsilon}{ct},
    \qquad
    T_t^v=-X_0+c\epsilon.
    \label{eq:linear_training_targets}
\end{equation}

\paragraph{Stochastic sampling}
We additionally study stochastic posterior sampling by varying the diffusion coefficient $w_t$ in~\eqref{eq:posterior_samplers}. Following \citet{ma2024sit}, we consider the KL-motivated diffusion coefficient
\begin{equation}
    w_t^{\mathrm{KL}}
    =
    2\left(
        \dot{\sigma}_t\sigma_t
        -
        \frac{\dot{\alpha}_t\sigma_t^2}{\alpha_t}
    \right),
    \label{eq:kl_diffusion_general}
\end{equation}
which minimizes an upper bound on the terminal KL divergence when numerical integration cost is disregarded. This choice also coincides with the variance rate of the linear Markov diffusion whose perturbation kernel has the same $(\alpha_t,\sigma_t)$. Consequently, $w_t=w_t^{\mathrm{KL}}$ recovers the corresponding conventional reverse-time SDE \citep{song2021scorebased}.

For the linear Gaussian path used in our experiments, $\alpha_t=1-t$ and $\sigma_t=ct$, the diffusion coefficient is:
\begin{equation}
    w_t^{\mathrm{KL}}
    =
    \frac{2c^2t}{1-t}.
    \label{eq:kl_diffusion_linear}
\end{equation}

Since we are free to use an arbitrary diffusion coefficient  \citep{albergo2023stochastic}, we additionally introduce a constant overall stochasticity level $\rho$:
\begin{equation}
    w_t
    =
    \rho\,w_t^{\mathrm{KL}},
    \qquad \rho\geq0,
    \label{eq:stochasticity_scale}
\end{equation}
where $\rho=0$ recovers deterministic sampling and $\rho=1$ recovers the KL-motivated reverse-SDE coefficient.

\subsection{Datasets and observation protocols}
\label{app:data_protocols}

\paragraph{Speech restoration.}
We use EARS \citep{richter24_ears} as the clean-speech dataset and consider three restoration tasks: speech enhancement, bandwidth extension, and \ac{STFT} phase retrieval. Speech enhancement follows the EARS-WHAM v2 protocol, while bandwidth extension and phase retrieval follow the task definitions of~\citet{welker2026realtime}. For the low-data experiments, we construct nested fixed subsets containing 1 and 10 minutes of paired data, corresponding to 6 and 60 ten-second pieces, respectively. The clean examples and their observations are generated once and kept fixed throughout training for all paired methods.

For bandwidth extension, observations remain at 16\,kHz but are low-pass filtered with a cutoff randomly selected between 2 and 4\,kHz, following the reference protocol. For phase retrieval, we use the \ac{STFT}-based observation defined by~\citet{welker2026realtime}. Hyperparameter selection uses the EARS validation split, and final evaluation uses the corresponding test split. We additionally evaluate speech enhancement out-of-domain on VoiceBank-DEMAND extended \footnote{https://www.emergentmind.com/topics/voicebank-demand-extended-vb-demandex} while training \ac{LSA} on EARS-WHAM pairs.

\paragraph{ImageNet-$256$.}
For ImageNet-$256$, we follow the public DAPS benchmark~\citep{zhang2025daps}. Evaluation uses the released subset of 100 ImageNet images \citep{imagenet}, and for each task we reproduce the corresponding observation using the forward operators and corruption parameters provided with the benchmark\footnote{https://github.com/zhangbingliang2019/DAPS}. We consider $4\times$ super-resolution, box and random inpainting, Gaussian, motion, and nonlinear deblurring, and high-dynamic-range reconstruction.

In the paired low-data setting, all methods use the same fixed subset of 128 ImageNet training images and one fixed observation per image. We additionally sample 25 non-intersecting ImageNet training images for validation. The clean training subset is shared across tasks, while observations are constructed separately for each degradation. Neither the degradation operator nor newly generated observations are available during low-data training.

\paragraph{ImageNet-$512$ $8\times$ super-resolution.}
For ImageNet-$512$, we follow the $8\times$ super-resolution protocol of \citet{xu2025rethinking}. We evaluate paired-data budgets of 25, 50, 100, and 1,000 training images. The 25-, 50-, and 100-image subsets are nested, while the 1,000-image subset is sampled independently. We reproduce the benchmark observation process and use the same 1,000-sample ImageNet validation set as the reference protocol.

\paragraph{Self-generated data setting.}
When the degradation operator is available, we additionally study a self-generated data setting. At each training epoch, we generate 128 clean samples from the frozen prior and add them to a growing training pool. Their observations are produced on the fly using the known degradation operator. Thus, no real paired training examples are used, although the operator is available during the training. Hyperparameters are selected on a separate set of 25 prior-generated validation samples and their degraded observations.

We use this setting for the known-operator experiments on ImageNet-$256$, ImageNet-$512$ and on EARS \ac{STFT} phase retrieval. We refer to it as the \emph{self-generated} setting.

\subsection{Models and training details}
\label{app:models_training}

\paragraph{Frozen priors.}
For speech, we train an unconditional NCSN++ model~\citep{song2021scorebased} from scratch on the full clean EARS training set using \eqref{eq:unconditional_fields} also known as Denoising Score Matching~\citep{song2021scorebased}. The model directly predicts the score and uses the standard $\sigma_t^2$ loss weighting. 
All speech representations use complex-\ac{STFT} magnitude compression $\widetilde{X}=\beta |X|^{\alpha}e^{j\angle X}$, where $|X|$ and $\angle X$ denote the \ac{STFT} magnitude and phase, respectively, $\alpha$ controls the magnitude compression, and $\beta$ is a global scaling factor.
Following the complex-\ac{STFT} speech-restoration SGMSE+ configuration \citep{richter2023speech}, we use  $\alpha=0.5$, $\beta=0.15$, and $\sigma_{\max}=0.5$ throughout all speech experiments.
The resulting prior contains 65.56M parameters and requires 132.70 GFLOPs per NFE for one second of audio.

For images, we use pretrained PixelDiT-XL models~\citep{yu2025pixeldit} at $256\times256$ and $512\times512$ resolution without further adaptation. We use the epoch-320 checkpoint for ImageNet-$256$ and the released ImageNet-$512$ checkpoint. The unconditional velocity field is obtained through the learned null class condition. Architectural configurations and compute per-step for the prior models are reported in Table~\ref{tab:prior_model_configurations}.

\paragraph{\ac{LSA} models and baselines.}
As the default architecture for all tasks, the \ac{LSA} is an NCSN++ model \citep{song2021scorebased} initialized from scratch. It receives $(x,y,t)$ as input and shares neither parameters nor intermediate features with the frozen prior. For speech, both the prior and \ac{LSA} are trained in score coordinates. For the main image experiments, the PixelDiT prior is used in its native velocity parameterization and \ac{LSA} is trained in the velocity parameterization. %
For an ablation study, we additionally train \ac{LSA} using a PixelDiT architecture or alternative score- and data-prediction parameterizations (see Appendix~\ref{app:field_conversion}). The corresponding results are reported in Tables~\ref{tab:lsa_parameterization_ablation} and~\ref{tab:lsa_backbone_ablation}.

For the PixelDiT baselines, we adapt ControlNet and LoRA to the original frozen PixelDiT backbone. Our ControlNet follows the transformer-based ControlNet-Half implementation of PixArt\footnote{\url{https://github.com/PixArt-alpha/PixArt-alpha/blob/master/diffusion/model/nets/pixart_controlnet.py}}. A trainable branch follows the first half of the patch-level transformer and injects residuals into the corresponding frozen PixelDiT blocks through zero-initialized projections. The observation image is embedded using the PixelDiT patch embedder, while the pixel-level blocks and output head remain unchanged. For our experiments, we use a width-reduced control branch ($d=692$) while preserving the ControlNet-Half topology.
LoRA~\citep{hu2022lora} uses the standard low-rank update and is applied to the joint QKV and output projections of all patch- and pixel-level attention blocks. The observation image is encoded by a two-layer conditioning stem and added to the patch tokens, and its final projection is zero-initialized. All original PixelDiT parameters remain frozen.

For the speech baselines, we adapt ControlNet~\citep{zhang2023controlnet} to the frozen NCSN++ score prior. The trainable ControlNet branch follows the encoder and middle block of NCSN++ and injects its features into the corresponding frozen backbone features through zero-initialized $1\times1$ convolutions. The observation $Y$ is represented by its real and imaginary \ac{STFT} components and provided to the trainable branch through a zero-initialized input projection. Consequently, the adapted model initially reproduces the frozen prior and gradually learns the observation-dependent correction during conditional score matching. For computational efficiency, we use a reduced-width ControlNet branch with $n_f=64$, compared with $n_f=128$ for the frozen prior.

For the DEFT~\citep{denker2024deft} baselines we use the authors'\footnote{https://github.com/alexdenker/DEFT} implementation and is train on our 128-pair DAPS subsets with the same image augmentations as the other paired methods. 
Unlike our main \ac{LSA} setting, the DEFT pipeline uses prior-model outputs and has an optional access to the degradation operator.
For speech, we also train a direct conditional posterior model with the same NCSN++ backbone on the corresponding 1- and 10-minute paired subsets as well as on the full training set.

All ControlNet, LoRA and direct posterior models trained on~\eqref{eq:conditional_fields} with velocity parameterization.

\paragraph{Optimization and augmentation.}
All \ac{LSA} models are trained using Adam with a learning rate of $2\times10^{-4}$ and batch size of 64.
We use a step learning rate scheduler, and reduce the learning rate to $2.5\times10^{-5}$ through 150 epoches. We use the final training checkpoint and do not select checkpoints based on a metric. To keep the effective number of optimization examples comparable across low-data regimes, small paired datasets are repeated within each training epoch; the underlying clean--observation pairs themselves remain fixed. Image models use the same pair-preserving dihedral augmentation for all methods, consisting of random rotations by multiples of $90^\circ$ and horizontal flipping. No additional observations are generated from the forward operator in the real-pair low-data setting.

\paragraph{Complexity reporting.}
All GFLOP counts in Tables~\ref{tab:prior_model_configurations}--\ref{tab:speech_model_configurations} are measured with \texttt{thop}\footnote{https://github.com/ultralytics/thop} and report the cost of one network function evaluation per image / 1 second of speech. The speech models accept variable-duration inputs.

\begin{table}[t]
\centering
\caption{
Prior model configurations used throughout the experiments. All prior parameters remain fixed during downstream training. Speech GFLOPs counts are measured for one second of audio.
}
\label{tab:prior_model_configurations}
\small
\setlength{\tabcolsep}{3.5pt}
\renewcommand{\arraystretch}{1.1}
\begin{tabularx}{\textwidth}{@{}>{\raggedright\arraybackslash}p{0.14\textwidth}>{\raggedright\arraybackslash}p{0.12\textwidth}cc>{\raggedright\arraybackslash}Xrr@{}}
\toprule
Setting
& Model
& Width
& \shortstack{Depth\\config.}
& Architecture details
& Frozen Parameters
& \shortstack[r]{GFLOPs/\\NFE} \\
\midrule
ImageNet-$256$ & PixelDiT-XL ($320$ ep.) & 1152 & 26 & 4 pixel blocks; patch size 16 & 797.38M & 311.19 \\
ImageNet-$512$ & PixelDiT-XL & 1152 & 26 & 4 pixel blocks; patch size 16 & 797.38M & 1352.24 \\
\midrule
\multirow{2}{*}{Speech} & NCSN++ & 128 & $(1,1,2,2,2,2,2)$ & 2 residual blocks & 65.56M & 132.70 \\
& ADM & 128 & $(1,1,2,2,2,2,2)$ & 2 residual blocks& 72.25M & 131.42 \\
\bottomrule
\end{tabularx}
\end{table}

\begin{table}[t]
\centering
\caption{
Model configurations used across all ImageNet experiments. Parameters and GFLOPs/NFE refer to the trainable \ac{LSA} model or adapter.
}
\label{tab:image_model_configurations}
\small
\setlength{\tabcolsep}{3.5pt}
\renewcommand{\arraystretch}{1.1}
\begin{tabularx}{\textwidth}{@{}>{\raggedright\arraybackslash}p{0.20\textwidth}>{\raggedright\arraybackslash}p{0.14\textwidth}cc>{\raggedright\arraybackslash}Xrr@{}}
\toprule
Setting
& Model
& Width
& \shortstack{Depth\\config.}
& Architecture details
& \shortstack[r]{Trainable\\params}
& \shortstack[r]{GFLOPs/\\NFE} \\
\midrule
IN256, 0 real pairs & LSA (NCSN++) & 40 & $(1,1,1,2)$ & 2 residual blocks & 2.16M & 44.30 \\
IN256, 128 pairs & LSA (NCSN++) & 32 & $(1,1,1,2)$ & 1 residual block & 1.01M & 19.84 \\
IN256, 128 pairs & ControlNet & 692 & 13 & 13 control blocks & 124.92M & 46.80 \\
IN256, 128 pairs & LoRA & -- & 30 & rank=8 & 2.15M & 1.10 \\
IN256 ablation & LSA (PixelDiT) & 96 & 4 & 2 pixel blocks; patch size 4 & 0.98M & 44.71 \\
\midrule
IN512, 0 real pairs & LSA (NCSN++) & 40 & $(1,1,1,2)$ & 2 residual blocks & 2.16M & 176.04 \\
\bottomrule
\end{tabularx}
\end{table}

\begin{table}[t]
\centering
\caption{
Speech model configurations. Parameters and GFLOPs/NFE refer to the trainable model shown in each row. Speech compute is measured for one second of audio.}
\label{tab:speech_model_configurations}
\small
\setlength{\tabcolsep}{3.5pt}
\renewcommand{\arraystretch}{1.1}
\begin{tabularx}{\textwidth}{@{}>{\raggedright\arraybackslash}p{0.14\textwidth}>{\raggedright\arraybackslash}p{0.18\textwidth}c>{\raggedright\arraybackslash}p{0.19\textwidth}>{\raggedright\arraybackslash}Xrr@{}}
\toprule
Setting
& Model
& Width
& Depth config.
& Architecture details
& \shortstack[r]{Trainable\\params}
& \shortstack[r]{GFLOPs/\\NFE} \\
\midrule
1 min & Conditional / LSA & 32 & $(1,2,2)$ & 2 residual blocks & 1.71M & 27.60 \\
10 min & Conditional / LSA & 64 & $(1,1,2,2)$ & 1 residual block & 5.47M & 46.51 \\
Full & Conditional / LSA & 128 & $(1,1,2,2,2,2,2)$ & 2 residual blocks & 65.59M & 266.03 \\
\begin{tabular}[t]{@{}l@{}}0 real pairs\\Phase Retrieval\end{tabular} & LSA & 64 & $(1,1,2,2)$ & 2 residual blocks & 7.60M & 66.22 \\
\midrule
1 / 10 min & ControlNet & 32 & 7/7 & -- & 1.81M & 5.55 \\
Full & ControlNet & 64 & 7/7 & -- & 6.66M & 19.64 \\
\bottomrule
\end{tabularx}
\end{table}

\subsection{Inference refinements}
\label{app:inference_refinements}

\paragraph{\ac{LSA} guidance.}
LSA is trained using the additive posterior-field estimator from~\eqref{eq:prior_residual_model}, corresponding to unit \ac{LSA} weight. At inference, we introduce a time-dependent \ac{LSA} weight
$\zeta_t$,
\begin{equation}
    \widehat f_{\zeta}(x,y,t)
    =
    f_\theta(x,t)
    +
    \zeta_t \widehat{\ell}_\psi^f(x,y,t).
    \label{eq:residual_scaling}
\end{equation}
We parameterize the guidance schedule by its values at the data and noise
endpoints, $\zeta_0$ and $\zeta_1$, and linearly interpolate between them,
\begin{equation}
    \zeta_t
    =
    (1-t)\zeta_0+t\zeta_1.
    \label{eq:linear_residual_guidance}
\end{equation}
Thus, $\zeta_0=\zeta_1=1$ recovers the field used during training.

\ac{LSA} scaling serves the same purpose as standard diffusion guidance~\citep{dhariwal2021diffusion,ho2022classifierfree}: it provides an inference-time control over the strength of conditioning. However, by~\eqref{eq:residual_optima}, \ac{LSA} contains both the likelihood score and the prior model error $\Delta_\theta^f$. Scaling it therefore does not correspond purely to likelihood tempering, and we interpret $\zeta_t$ more generally as a guidance-like control of the learned posterior field.

\paragraph{Restart refinement.}
We additionally use a short restart-style refinement procedure motivated by restart sampling and posterior restart methods~\citep{xu2023restart,zhang2025daps,ahmed2025reps}.
After completing a reverse trajectory and obtaining an estimate $\widehat X_0$, we perturb it to an intermediate time $\tau$ of the same Gaussian path,
\begin{equation}
    X_\tau
    =
    \alpha_\tau \widehat X_0
    +
    \sigma_\tau \epsilon,
    \qquad
    \epsilon\sim\mathcal N(0,I),
    \label{eq:jump_refinement}
\end{equation}
and integrate the conditional trajectory again from $\tau$ toward the data endpoint using the original observation $y$. The procedure can be repeated for multiple refinement rounds. The jump time and number of rounds are selected on held-out validation data, and all additional field evaluations are included in the reported NFE. The impact of both inference refinements is ablated in Table~\ref{tab:speech_refinement_ablation}.

\paragraph{Hyperparameter optimization.}
Inference hyperparameters are optimized on held-out validation data using Ray Tune with the HyperOpt search algorithm.\footnote{https://docs.ray.io/en/latest/tune/api/doc/ray.tune.search.hyperopt.HyperOptSearch.html}
The search jointly considers the number of integration steps, the endpoint \ac{LSA} weights $(\zeta_0,\zeta_1)$, and the restart-refinement parameters when enabled. In our current search space, NFE is varied over integer values from 5 to 30, while $\zeta_0$ and $\zeta_1$ are selected from $[0.0, 5.0]$ in increments of  $0.25$. The restart time $\tau$ is searched within a range of $[0.2, 0.8]$ and by default we use a single restart.

For stochastic sampling and score-parameterized models, we additionally optimize the sampling start time $t_{\mathrm{start}}\in[0.8,0.999]$ to avoid the endpoint singularity at $t=1$. We also select between initialization from $\mathcal{N}(0,\sigma_{t_{\mathrm{start}}}^2I)$ and an observation-centered Gaussian $\mathcal{N}(\alpha_{t_{\mathrm{start}}}y,\sigma_{t_{\mathrm{start}}}^2I)$.

We run 150 HyperOpt trials for each task and method. The five best configurations found by the search are subsequently reevaluated on the full validation set, from which the final configuration is selected. The selected configuration is then kept fixed for test evaluation.

For speech enhancement and STFT phase retrieval optimization maximizes PESQ, whereas for bandwidth extension it minimizes LSD.
For ImageNet experiments, optimization minimizes LPIPS on 25 held-out training images; in the self-generated data setting, these are replaced by 25 independently generated prior samples and their corresponding degraded observations. The same validation-based optimization procedure is applied to LSA, ControlNet, LoRA, and the direct posterior baseline over the sampling hyperparameters applicable to each method.

\raggedbottom
\section{ImageNet-$256$ results}
\label{app:imagenet256_low_data_results}

\subsection{Low-Data ImageNet-$256$ results}
\input{tables/image_low_data_results.tex}

\clearpage
\begin{figure}[H]
    \centering
    \includegraphics[width=\textwidth,height=0.88\textheight,keepaspectratio]{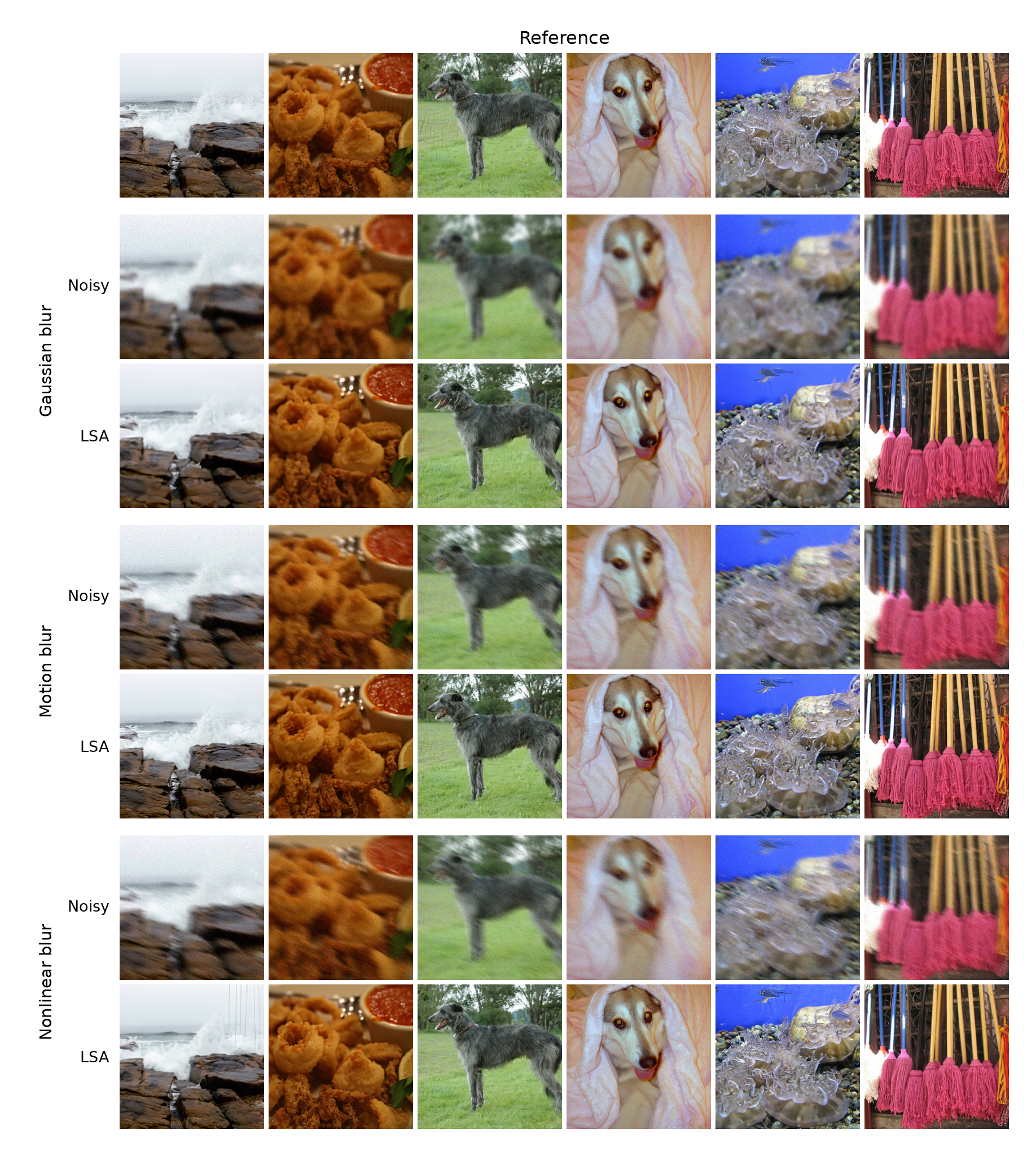}
    \caption{ImageNet-256 blur restoration: references, observations, and \ours{} reconstructions.}
    \label{fig:imagenet256_qualitative_blurs}
\end{figure}

\clearpage
\begin{figure}[H]
    \centering
    \includegraphics[width=\textwidth,height=0.88\textheight,keepaspectratio]{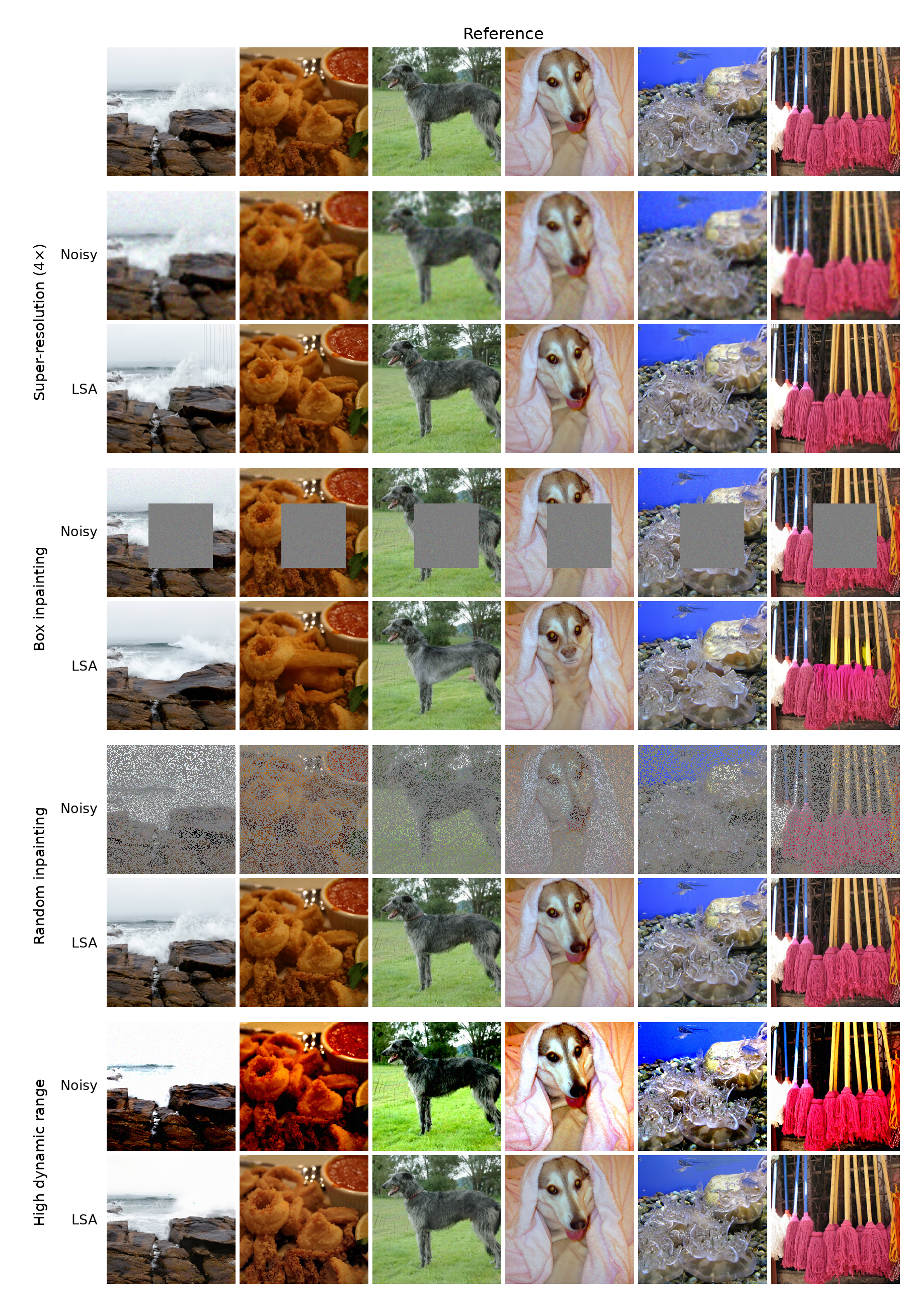}
    \caption{ImageNet-256 super-resolution, inpainting, and HDR restoration results.}
    \label{fig:imagenet256_qualitative_other_operators}
\end{figure}

\clearpage

\section{ImageNet-$512$ super-resolution results}
\label{app:imagenet512_srx8_results}

\input{tables/imagenet512_srx8_low_data.tex}

\section{Gaussian deblurring ablation}
\label{app:gaussian_blur_ablation}
\begin{figure}[H]
    \centering
    \includegraphics[width=\textwidth]{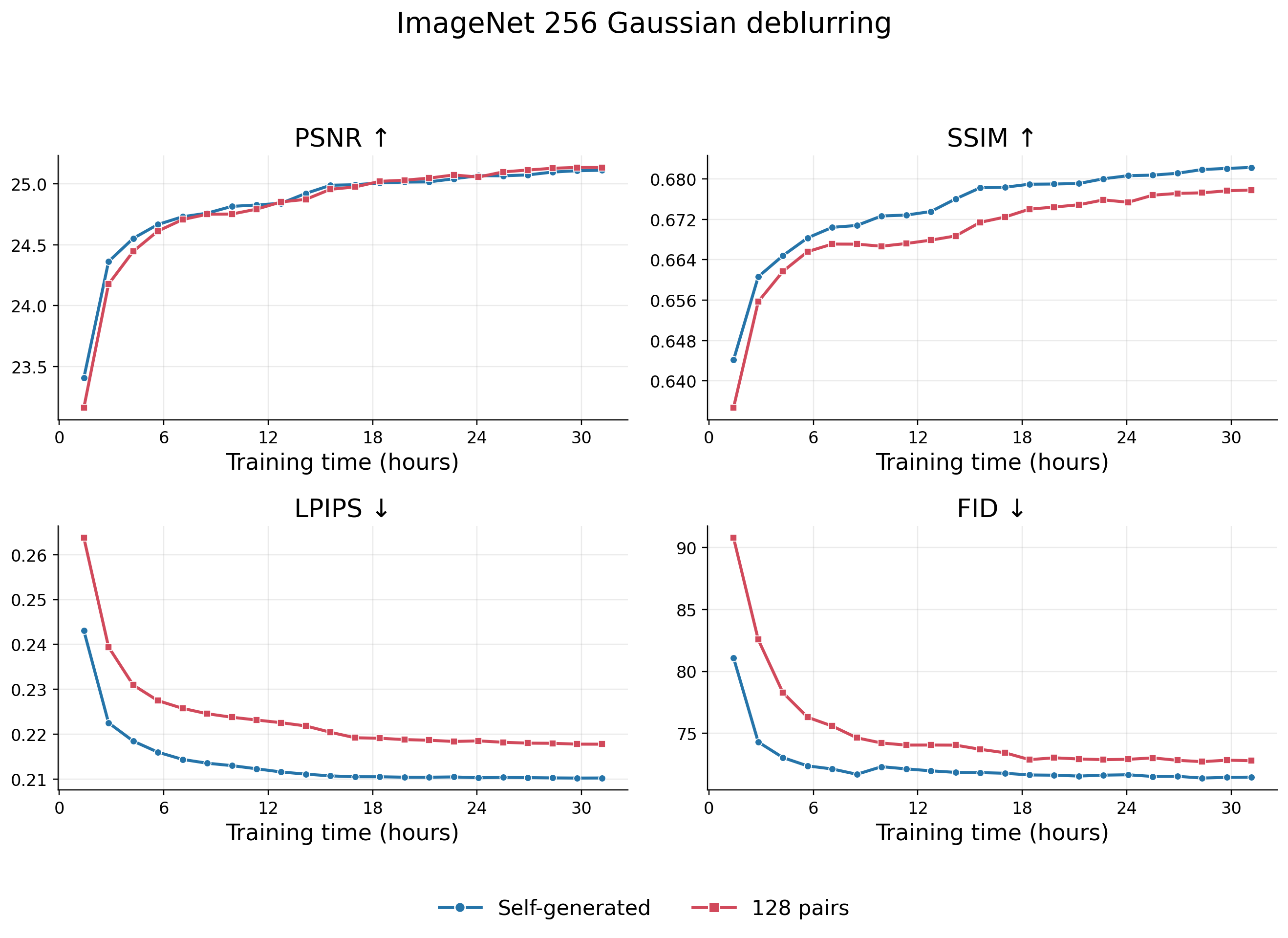}
    \caption{Gaussian-deblurring learning curves for \ac{LSA} on ImageNet-$256$ with a PixelDiT prior, trained using either 128 real pairs or self-generated data. Training time is measured on a single A100 GPU.
    }
    \label{fig:gaussian_blur_learning_curves}
\end{figure}

Tables~\ref{tab:lsa_parameterization_ablation} and \ref{tab:lsa_backbone_ablation} report the parameterization and architecture ablations, respectively, while Figure~\ref{fig:gaussian_blur_learning_curves} shows the corresponding training dynamics.

\input{tables/gaussian_blur_parameterization_ablation.tex}

\input{tables/gaussian_blur_backbone_ablation.tex}

\clearpage
\section{Bandwidth extension qualitative results}
\label{app:bwe_qualitative_ablation}

\begin{figure}[H]
    \centering
    \includegraphics[width=\textwidth,height=0.82\textheight,keepaspectratio]{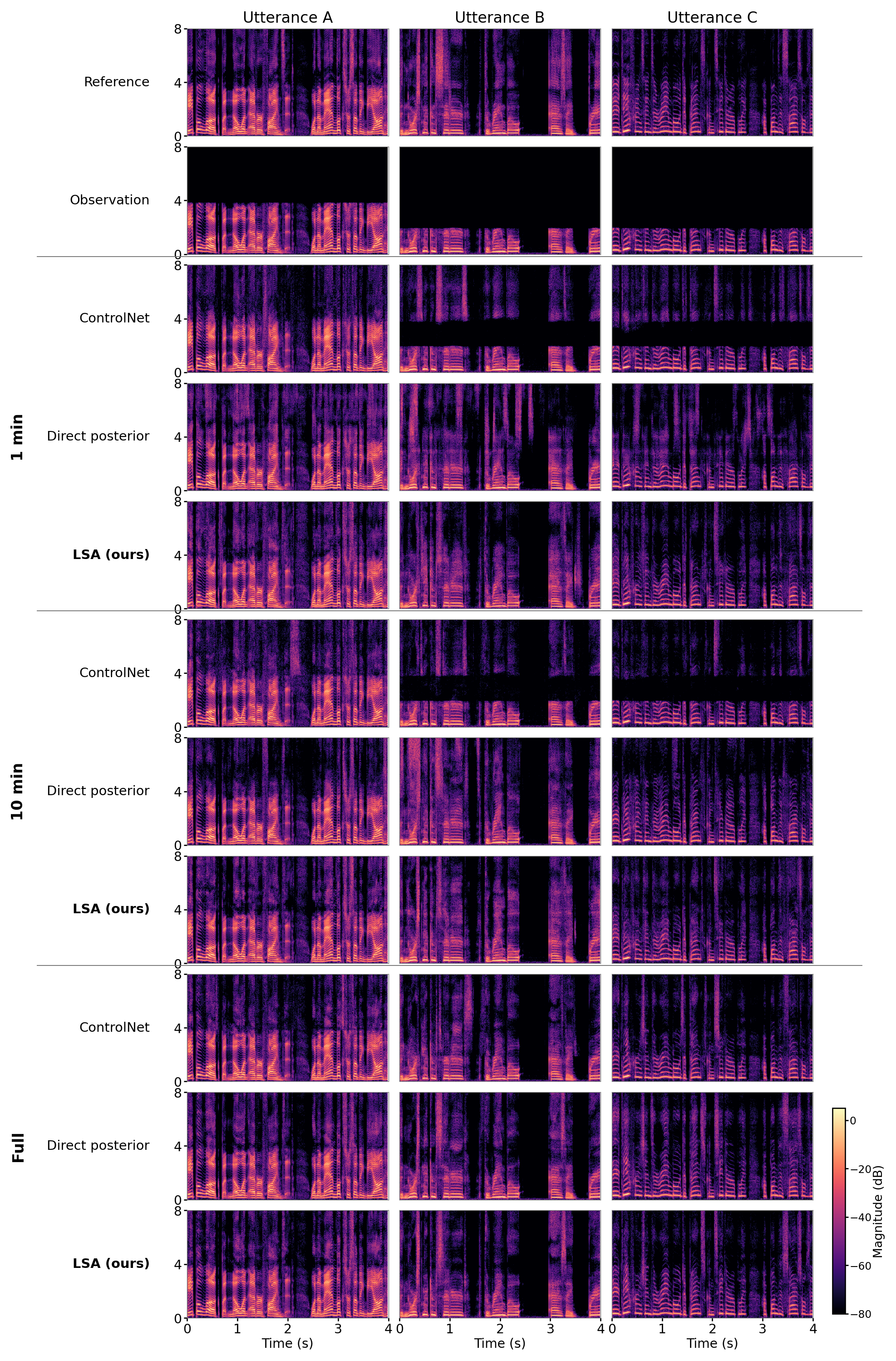}
    \caption{Qualitative BWE comparison on three EARS test utterances. The shared top rows show the clean reference and low-pass-filtered observation; each subsequent block compares ControlNet, direct posterior training, and \ours{} using 1 minute, 10 minutes, or the full paired training set. Spectrogram magnitudes are measured relative to the clean-reference peak of each utterance, with frequency shown in kHz.}
    \label{fig:bwe_qualitative_data_scaling}
\end{figure}

\section{Prior-swapping ablation}
\label{app:prior_swapping}

\begin{figure}[H]
    \centering    \includegraphics[width=0.7\textwidth,height=0.88\textheight,keepaspectratio]{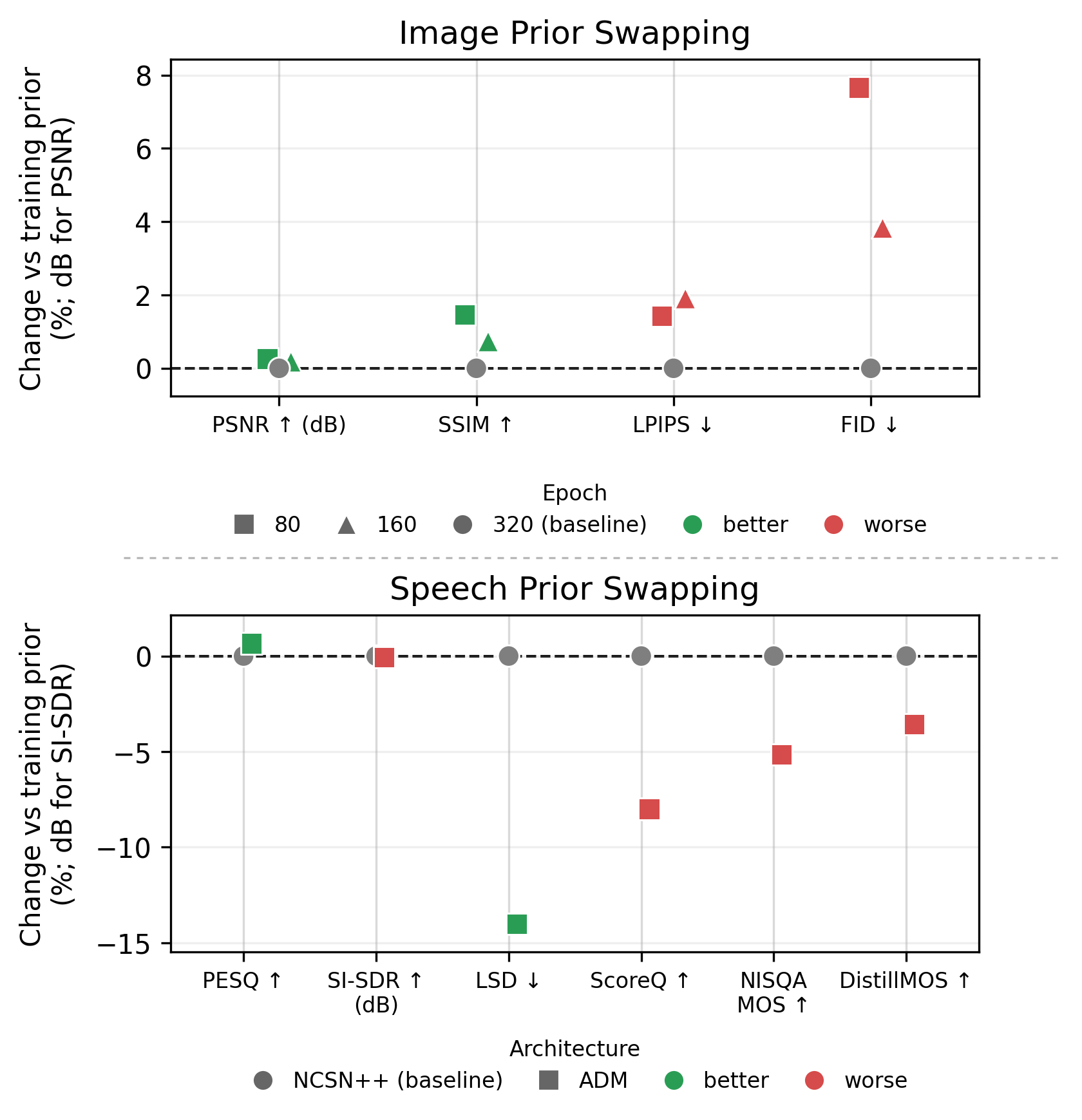}
    \caption{
    Inference-time prior swapping without retraining \ac{LSA}. 
    Top: ImageNet-$256$ Gaussian deblurring with 128 pairs, where the PixelDiT prior used during training (epoch 320) is replaced by earlier checkpoints. 
    Bottom: 1-min speech-enhancement \ac{LSA}, where the NCSN++ training prior is replaced by an ADM prior trained on the same EARS data. 
    Values show metric changes relative to the training prior; despite the architecture change, performance remains broadly comparable, with a substantial improvement in LSD.
    }
    \label{fig:prior_swapping}
\end{figure}

\section{Stochastic-sampling ablation}
\label{app:stochastic_sampling}

\begin{figure}[H]
    \centering
    \includegraphics[width=0.8\textwidth]{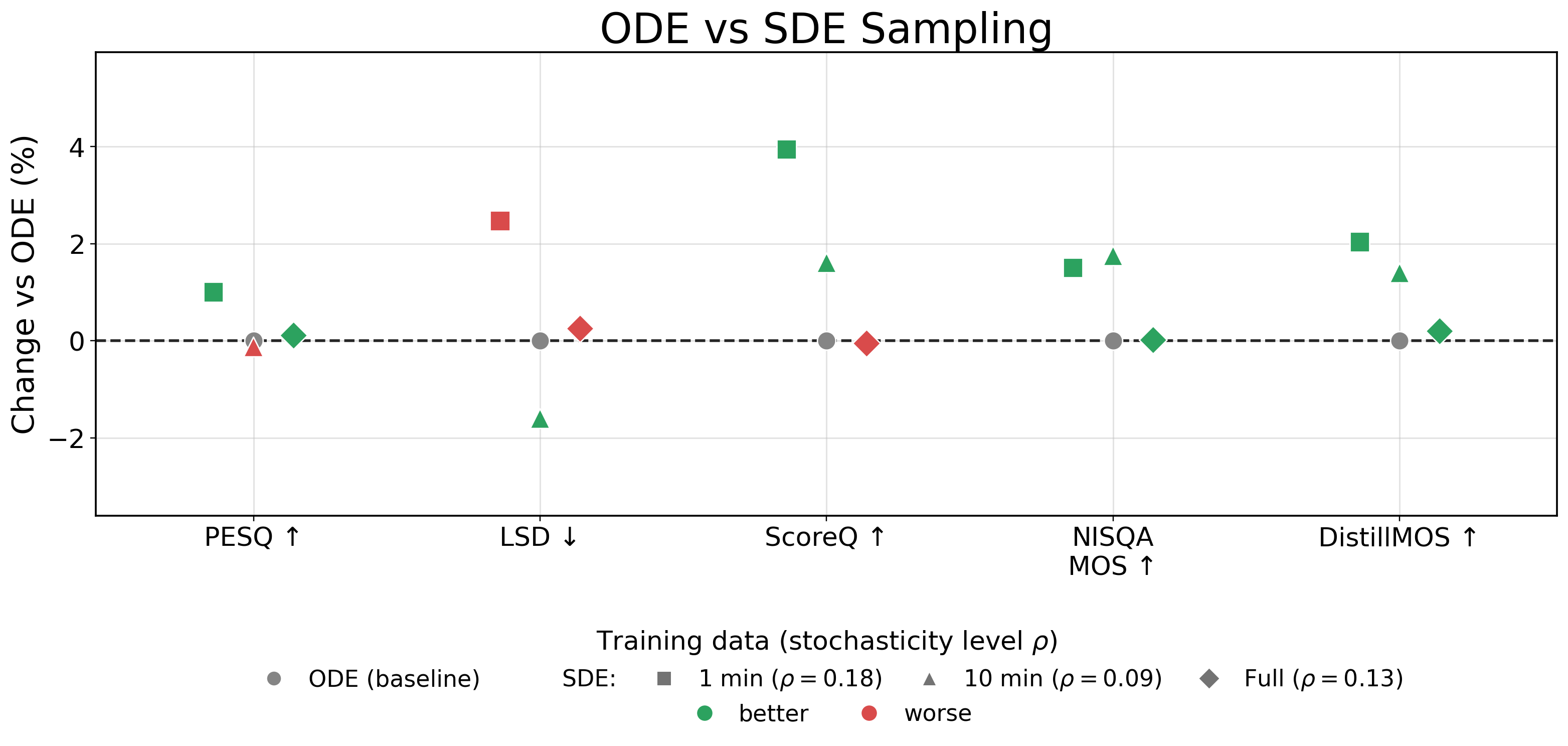}
    \caption{ODE versus SDE sampling for \ac{LSA} bandwidth extension with 1 min, 10 min, and full paired data. Changes are relative to the ODE baseline ($\rho=0$); SDE uses $\rho=0.18$, $0.09$, and $0.13$, respectively. Mild stochasticity generally improves performance, providing complementary gains.}
    \label{fig:bwe_ode_vs_sde}
\end{figure}

\section{Inference refinement ablation}

\begin{table}[H]
\centering
\caption{
Inference-refinement ablation for the 1-min \ac{LSA} speech-enhancement model. We vary the use of restart refinement and the \ac{LSA} guidance-weight schedule $\zeta_1{\rightarrow}\zeta_0$, while keeping the trained model fixed (see Appendix~\ref{app:inference_refinements}).
}
\label{tab:speech_refinement_ablation}
\small
\setlength{\tabcolsep}{4pt}
\renewcommand{\arraystretch}{1.08}
\begin{tabular}{@{}lccccc@{}}
\toprule
Restart
& $\zeta_1{\rightarrow}\zeta_0$
& PESQ $\uparrow$
& SI-SDR $\uparrow$
& Avg.\ nMOS $\uparrow$
& FAD $\downarrow$ \\
\midrule
Yes & $3{\rightarrow}0$
& 1.51 & 7.82 & \textbf{3.81} & \textbf{0.35} \\
No & $3{\rightarrow}0$
& \textbf{1.52} & 7.67 & 3.47 & 1.29 \\
Yes & $1$
& 1.47 & 7.43 & 3.14 & 0.70 \\
Yes & $2$
& 1.51 & \textbf{10.56} & 2.93 & 1.82 \\
No & $1$
& 1.36 & 5.96 & 2.76 & 1.11 \\
\bottomrule
\end{tabular}
\end{table}

\end{document}

%% file: math_commands.tex
\usepackage{amsmath,amsfonts,bm}

\def\secref#1{section~\ref{#1}}

\def\eqref#1{equation~\ref{#1}}

\def\1{\bm{1}}

\DeclareMathAlphabet{\mathsfit}{\encodingdefault}{\sfdefault}{m}{sl}
\SetMathAlphabet{\mathsfit}{bold}{\encodingdefault}{\sfdefault}{bx}{n}

%% file: tables/speech_results_compact.tex
\begin{table}[t]
    \centering
     \caption{Speech restoration results. (E), (U), and (VB) denote EARS, Urgent, and VoiceBank priors. $^{*}$AnyEnhance is trained on 44.1-kHz audio. \textbf{Bold}: best within each data regime.}
    \label{tab:speech_main_results}
    \label{tab:bwe_main_results}
    \label{tab:phase_retrieval_results}
    \scriptsize
    \setlength{\tabcolsep}{2.2pt}
    \begin{minipage}[t]{0.52\textwidth}
        \centering
        \textbf{(a) Speech enhancement}\par\smallskip
        \resizebox{\linewidth}{!}{%
        \begin{tabular}{clcccc}
            \toprule
            Data & Model & PESQ $\uparrow$ & SI-SDR $\uparrow$ & Avg. nMOS $\uparrow$ & FAD $\downarrow$ \\
            \midrule
            \multicolumn{6}{l}{\textit{EARS-WHAM-v2}} \\
            \cmidrule(lr){1-6}
            \multirow{3}{*}{1 min}
                & Direct Posterior  & 1.44 & 8.70 & 2.60 & 1.95 \\
                & ControlNet & 1.48 & 9.66 & 3.33 & 0.74 \\
                & \ours     & \textbf{1.63} & \textbf{9.80} & \textbf{3.60} & \textbf{0.58} \\
            \cmidrule(lr){1-6}
            \multirow{3}{*}{10 min}
                & Direct Posterior  & 1.70 & 11.50 & 3.44 & 0.25 \\
                & ControlNet & 1.47 & 10.18 & 3.43 & 0.42 \\
                & \ours     & \textbf{1.86} & \textbf{11.90} & \textbf{3.94} & \textbf{0.24} \\
            \cmidrule(lr){1-6}
            \multirow{5}{*}{Full}
                & SBVE                 & 2.15 & 12.90 & 4.09 & 1.05 \\
                \cmidrule(lr){2-6}
                & Direct Posterior            & 2.14 & \textbf{14.75} & 3.88 & 0.54 \\
                & ControlNet & 1.63 & 12.62 & 3.49 & 0.46 \\
                & \ours{} (E) & 2.21 & 14.35 & \textbf{4.17} & \textbf{0.30} \\
                & \ours{} (U) & \textbf{2.24} & 14.16 & 4.06 & 0.50 \\
            \midrule
            \multicolumn{6}{l}{\textit{VoiceBank Demand-ex (out-of-domain evaluation)}} \\
            \cmidrule(lr){1-6}
            \multirow{4}{*}{Full}
                & Direct Posterior                 & 2.23 & \textbf{13.22} & 3.94 & 0.89 \\
                & \ours{} (E)  & 2.19 & 9.10  & 4.01 & 1.02 \\
                & \ours{} (U)  & \textbf{2.32} & 11.25 & \textbf{4.07} & 0.78 \\
                & \ours{} (VB) & \textbf{2.32} & 9.87  & 4.05 & \textbf{0.52} \\
            \bottomrule
        \end{tabular}%
        }
    \end{minipage}\hfill
    \begin{minipage}[t]{0.46\textwidth}
        \centering
        \textbf{(b) Bandwidth extension}\par\smallskip
        \resizebox{\linewidth}{!}{%
        \input{tables/bwe_main_results_v2.tex}        }

        \vspace{1.0em}
        \textbf{(c) STFT phase retrieval}\par\smallskip
        \resizebox{\linewidth}{!}{%
        \begin{tabular}{clcccc}
            \toprule
            Data & Model & PESQ $\uparrow$ & LSD $\downarrow$ & Avg. nMOS $\uparrow$ & FAD $\downarrow$ \\
            \midrule
            \multirow{2}{*}{1 min}
                & ControlNet & \textbf{4.34} & 0.81 & {4.07} & 0.079 \\
                & \ours{} & 4.26 & \textbf{0.42} & 4.07 & \textbf{0.059} \\
            \cmidrule(lr){1-6}
            \multirow{2}{*}{self-gen.}
                & ControlNet & 4.37 & 0.70 & 4.09 & 0.062 \\
                & \ours{} & \textbf{4.55} & \textbf{0.28} & \textbf{4.19} & \textbf{0.011} \\
            \cmidrule(lr){1-6}
            \multirow{2}{*}{Full}
                & DiffPhase & 4.04 & 0.53 & 4.10 & 0.046 \\
                & StreamFlow & 4.24 & 0.76 & 4.20 & 0.106 \\
            \bottomrule
        \end{tabular}%
        }
    \end{minipage}
\end{table}

%% file: tables/bwe_main_results_v2.tex
\begin{tabular}{clcccc}
    \toprule
    Data & Model & PESQ $\uparrow$ & LSD $\downarrow$ & Avg. nMOS $\uparrow$ & FAD $\downarrow$ \\
    \midrule
    \multirow{3}{*}{1 min}
        & Direct Posterior & 1.81 & 1.40 & 3.43 & 0.38 \\
        & ControlNet & 2.21 & 1.39 & 3.48 & 1.53 \\
        & \ours & \textbf{2.24} & \textbf{1.20} & \textbf{3.79} & \textbf{0.32} \\
    \cmidrule(lr){1-6}
    \multirow{3}{*}{10 min}
        & Direct Posterior & 2.42 & 1.11 & 3.86 & 0.20 \\
        & ControlNet & 2.20 & 1.43 & 3.44 & 1.40 \\
        & \ours & \textbf{2.53} & \textbf{1.09} & \textbf{3.98} & \textbf{0.19} \\
    \cmidrule(lr){1-6}
    \multirow{5}{*}{Full}
        & AnyEnhance$^{*}$ & 2.50 & 1.40 & 4.12 & 0.64 \\
        & StreamFlow & \textbf{3.37} & 1.26 & 3.78 & 0.71 \\
    \cmidrule(lr){2-6}
        & Direct Posterior & 2.91 & 1.03 & 4.18 & \textbf{0.11} \\
        & ControlNet & 2.27 & 1.20 & 3.83 & 0.49 \\
        & \ours & 2.90 & \textbf{0.99} & \textbf{4.20} & 0.15 \\
    \bottomrule
\end{tabular}

%% file: tables/image_no_real_pairs_results.tex
\begin{table}[t]
    \caption{ImageNet-$256$ restoration on 100 images with additive Gaussian noise ($\sigma_y=0.05$), following the DAPS benchmark~\citep{zhang2025daps}. \textbf{Bold}: best. $^{\dagger}$DAPS rerun from DAPS++~\citep{chen2025dapspp}.}
    \label{tab:image_no_real_pairs_results}
    \centering
    \scriptsize
    \setlength{\tabcolsep}{3.0pt}
    \renewcommand{\arraystretch}{1.1}
    \setlength{\aboverulesep}{0pt}
    \setlength{\belowrulesep}{0pt}
    \resizebox{\textwidth}{!}{%
    \begin{tabular}{llc!{{\color{black!50}\vrule width 0.4pt}}cccc!{\hspace{3pt}\vrule width 0.6pt\hspace{3pt}}llc!{{\color{black!50}\vrule width 0.4pt}}cccc}
        \toprule\addlinespace[0.65ex]
        Task & Method & NFE
        & PSNR $\uparrow$ & SSIM $\uparrow$ & LPIPS $\downarrow$
        & FID $\downarrow$
        & Task & Method & NFE
        & PSNR $\uparrow$ & SSIM $\uparrow$ & LPIPS $\downarrow$
        & FID $\downarrow$ \\
        \addlinespace[0.4ex]\midrule\addlinespace[0.65ex]
        \multirow[c]{4}{*}{\begin{tabular}[t]{@{}l@{}}Super-\\resolution\\($4\times$)\end{tabular}}
        & DMAP & $\sim$300 & 25.39 & .661 & .229 & 74.65
        & \multirow[c]{3}{*}{\begin{tabular}[t]{@{}l@{}}Gaussian\\deblurring\end{tabular}}
        & DAPS & 1k & \textbf{26.15} & .684 & .253 & 75.68 \\
        & DAPS & 1k & \textbf{25.89} & .694 & .276 & 83.57
        & & ControlNet & 26 & 25.08 & .668 & .242 & 99.82 \\
        & ControlNet & 26 & 25.06 & .675 & .237 & 92.15
        & & \ours{} & 21 & 25.39 & \textbf{.690} & \textbf{.212} & \textbf{69.34} \\
        \cmidrule(lr){8-14}
        & \ours{} & 23 & 25.57 & \textbf{.700} & \textbf{.215} & \textbf{70.38}
        & \multirow[c]{3}{*}{\begin{tabular}[t]{@{}l@{}}Motion\\deblurring\end{tabular}}
        & RePS & 1k & \textbf{28.95} & .801 & .169 & 53.15 \\
        \cmidrule(lr){1-7}
        \multirow[c]{4}{*}{\begin{tabular}[t]{@{}l@{}}Inpainting\\(box)\end{tabular}}
        & MAS & -- & 21.15 & \textbf{.817} & .168 & 95.96
        & & DAPS$^{\dagger}$ & 1k & -- & .769 & .175 & 47.09 \\
        & DAPS & 1k & \textbf{21.43} & .725 & .214 & 109.85
        & & \ours{} & 42 & 28.18 & \textbf{.802} & \textbf{.130} & \textbf{32.45} \\
        \cmidrule(lr){8-14}
        & ControlNet & 14 & 20.59 & .683 & .267 & 149.97
        & \multirow[c]{3}{*}{\begin{tabular}[t]{@{}l@{}}Nonlinear\\deblurring\end{tabular}}
        & RED-diff & 1k & \textbf{30.07} & .754 & .211 & 51.22 \\
        & \ours{} & 34 & 20.42 & .782 & \textbf{.161} & \textbf{92.43}
        & & DAPS & 4k & 27.73 & .724 & .169 & 59.87 \\
        \cmidrule(lr){1-7}
        \multirow[c]{3}{*}{\begin{tabular}[t]{@{}l@{}}Inpainting\\(random)\end{tabular}}
        & RePS & 1k & \textbf{28.97} & .829 & .115 & 27.52
        & & \ours{} & 18 & 27.10 & \textbf{.789} & \textbf{.148} & \textbf{41.78} \\
        \cmidrule(lr){8-14}
        & DAPS & 1k & 28.44 & .775 & .135 & 54.25
        & \multirow[c]{3}{*}{\begin{tabular}[t]{@{}l@{}}High\\dynamic\\range\end{tabular}}
        & RePS & 4k & 26.37 & .843 & .157 & 37.23 \\
        & \ours{} & 12 & 28.75 & \textbf{.847} & \textbf{.113} & \textbf{27.36}
        & & DAPS & 4k & 26.30 & .717 & .175 & 64.19 \\
        & & & & & &
        & & \ours{} & 20 & \textbf{28.30} & \textbf{.871} & \textbf{.120} & \textbf{25.64} \\

        \addlinespace[0.4ex]\bottomrule
    \end{tabular}%
    }
\end{table}

%% file: tables/image_low_data_results.tex
\begin{table}[H]
    \caption{Detailed ImageNet-$256$ restoration results using 128 real pairs.
    We report NFE, PSNR, SSIM, LPIPS, and FID across all evaluated tasks.
    Bold denotes the best result for each task.}
    \label{tab:image_low_data_results}
    \centering
    \small
    \setlength{\tabcolsep}{6.0pt}
    \begin{tabular}{llc!{{\color{black!50}\vrule width 0.4pt}}cccc}
        \toprule
        Task & Method & NFE
        & PSNR $\uparrow$ & SSIM $\uparrow$ & LPIPS $\downarrow$
        & FID $\downarrow$ \\
        \midrule
        \multirow[t]{3}{*}{Super-resolution ($4\times$)}
        & DEFT & 100 & 24.87 & .651 & .244 & 97.15 \\
        & ControlNet & 43 & 24.78 & .665 & .247 & 89.14 \\
        & \ours{} & 21 & \textbf{24.93} & \textbf{.683} & \textbf{.229} & \textbf{73.01} \\
        \midrule
        \multirow[t]{3}{*}{Inpainting (box)}
        & DEFT & 100 & 19.20 & \textbf{.759} & .194 & 141.19 \\
        & ControlNet & 10 & \textbf{20.24} & .685 & .275 & 152.02 \\
        & \ours{} & 22 & 18.94 & .755 & \textbf{.193} & \textbf{134.00} \\
        \midrule
        \multirow[t]{3}{*}{Inpainting (random)}
        & DEFT & 100 & 27.00 & .805 & .156 & 41.70 \\
        & ControlNet & 20 & 23.48 & .632 & .290 & 112.99 \\
        & \ours{} & 24 & \textbf{27.28} & \textbf{.808} & \textbf{.127} & \textbf{32.77} \\
        \midrule
        \multirow[t]{3}{*}{Gaussian deblurring}
        & ControlNet & 42 & \textbf{24.71} & .650 & .262 & 106.45 \\
        & LoRA & 36 & 24.28 & .642 & .259 & 106.18 \\
        & \ours{} & 36 & \textbf{24.86} & \textbf{.661} & \textbf{.221} & \textbf{73.15} \\
        \midrule
        \multirow[t]{3}{*}{Nonlinear deblurring}
        & ControlNet & 55 & 23.35 & .621 & .278 & 119.78 \\
        & LoRA & 20 & 23.93 & .657 & .255 & 100.87 \\
        & \ours{} & 21 & \textbf{24.59} & \textbf{.731} & \textbf{.185} & \textbf{52.21} \\
        \bottomrule
    \end{tabular}%
\end{table}

%% file: tables/imagenet512_srx8_low_data.tex
    \begin{table}[H]
        \vspace{0pt}
        \centering
        \caption{ImageNet-$512$ $8\times$ super-resolution across paired-data budgets. 
        LSA is trained with 25, 50, 100, or 1k real pairs; ``self-gen'' denotes training on self-generated pairs obtained using the known degradation operator. 
        Baseline results are from \citet{xu2025rethinking}.
        Bold indicates the best available value within each data budget.}
        \label{tab:imagenet512_srx8_low_data}
        \small
        \setlength{\tabcolsep}{6.0pt}
        \begin{tabular}{lcc!{{\color{black!50}\vrule width 0.4pt}}ccc}
            \toprule
            Source & Files & NFE $\downarrow$ & PSNR $\uparrow$
            & LPIPS $\downarrow$ & FID $\downarrow$ \\
            \midrule
            \rowcolor{black!6}
            DPS+CSE & 25 & 500 & 22.26
            & .405 & 75.64 \\
            \ours{} & 25 & 12 & \textbf{23.07}
            & \textbf{.347} & \textbf{40.76} \\
            \midrule
            \rowcolor{black!6}
            DPS+CSE & 50 & 500 & 22.84
            & .353 & 55.78 \\
            \ours{} & 50 & 9 & \textbf{23.20}
            & \textbf{.335} & \textbf{39.34} \\
            \midrule
            CSE & 100 & 500 & 16.54
            & .484 & 85.58 \\
            \rowcolor{black!6}
            DPS+CSE & 100 & 500 & 22.98
            & .323 & 44.59 \\
            DMAP+CSE & 100 & 500 & \textbf{23.58}
            & .336 & 39.56 \\
            \ours{} & 100 & 20 & 23.32
            & \textbf{.321} & \textbf{36.48} \\
            \midrule
            CSE & 1k & 500 & 16.68
            & .466 & 67.38 \\
            \rowcolor{black!6}
            DPS+CSE & 1k & 500 & 23.35
            & .314 & 36.23 \\
            DMAP+CSE & 1k & 500 & \textbf{23.71}
            & .325 & 36.64 \\
            \ours{} & 1k & 20 & 23.26
            & \textbf{.312} & \textbf{33.55} \\
            \midrule
            CSE & self-gen & 500 & 17.15
            & .479 & 82.32 \\
            \rowcolor{black!6}
            DPS+CSE & self-gen & 500 & 23.09
            & .308 & 38.60 \\
            DMAP+CSE & self-gen & 500 & \textbf{23.33}
            & .311 & 38.29 \\
            \ours{} & self-gen & 32 & 23.11
            & \textbf{.297} & \textbf{28.90} \\
            \bottomrule
        \end{tabular}%
    \end{table}

%% file: tables/gaussian_blur_parameterization_ablation.tex
\begin{table}[H]
    \centering
    \caption{LSA parameterization ablation for Gaussian deblurring on ImageNet-$256$ with 128 pairs and an NCSN++ backbone. For each parameterization, we report the validation-selected sampling start time $t_{\mathrm{start}}$ and initialization. For velocity and data prediction, we additionally report default sampling from the Gaussian endpoint at $t_{\mathrm{start}}=1$; this setting is not defined for the score parameterization due to the endpoint singularity.}
    \label{tab:lsa_parameterization_ablation}
    \small
    \setlength{\tabcolsep}{2.8pt}
    \renewcommand{\arraystretch}{1.1}
    \begin{tabular}{@{}llrr!{{\color{black!50}\vrule width 0.4pt}}rrrr@{}}
        \toprule
        Parameterization & Initialization
        & $t_{\mathrm{start}}$ & NFE & PSNR $\uparrow$ & SSIM $\uparrow$ & LPIPS $\downarrow$ & FID $\downarrow$ \\
        \midrule
        Score & Observation marginal & .878 & 41 & 24.91 & .663 & .217 & 72.86 \\
        \midrule
        Velocity & Gaussian & .907 & 49 & 24.68 & .657 & .216 & 70.23 \\
        Velocity & Gaussian & 1.000 & 36 & 24.86 & .661 & .221 & 73.15 \\
        \midrule
        Data prediction & Observation marginal & .819 & 31 & 25.10 & .676 & .229 & 75.72 \\
        Data prediction & Gaussian & 1.000 & 29 & 24.49 & .655 & .233 & 74.19 \\
        \bottomrule
    \end{tabular}
\end{table}

%% file: tables/gaussian_blur_backbone_ablation.tex
\begin{table}[H]
    \centering
    \caption{LSA backbone ablation for Gaussian deblurring on ImageNet-$256$ with 128 pairs. We compare NCSN++ and PixelDiT using Gaussian initialization, reporting both the validation-selected sampling start time $t_{\mathrm{start}}$ and sampling from the Gaussian endpoint at $t_{\mathrm{start}}=1$.
    (Appendix~\ref{app:inference_refinements})
    }
    \label{tab:lsa_backbone_ablation}
    \small
    \setlength{\tabcolsep}{3.5pt}
    \renewcommand{\arraystretch}{1.1}
    \begin{tabular}{@{}llrr!{{\color{black!50}\vrule width 0.4pt}}rrrr@{}}
        \toprule
        FM backbone & Initialization
        & $t_{\mathrm{start}}$ & NFE & PSNR $\uparrow$ & SSIM $\uparrow$ & LPIPS $\downarrow$ & FID $\downarrow$ \\
        \midrule
        NCSN++ & Gaussian & .907 & 49 & 24.68 & .657 & .216 & 70.23 \\
        NCSN++ & Gaussian & 1.000 & 36 & 24.86 & .661 & .221 & 73.15 \\
        \midrule
        PixelDiT & Gaussian & .840 & 22 & 25.25 & .682 & .213 & 68.31 \\
        PixelDiT & Gaussian & 1.000 & 19 & 24.97 & .673 & .217 & 71.52 \\
        \bottomrule
    \end{tabular}
\end{table}